\documentclass[acmsmall]{acmart}
\usepackage{wrapfig}

\AtBeginDocument{%
  }

\setcopyright{acmlicensed}
\copyrightyear{2024}
\acmYear{2024}
\acmDOI{XXXXXXX.XXXXXXX}

\acmJournal{JACM}
\acmVolume{37}
\acmNumber{4}
\acmArticle{1}
\acmMonth{8}

\usepackage{url}
\usepackage{times}
\usepackage{epsfig}
\usepackage{graphicx}

\usepackage{amsmath}

\usepackage{amssymb}
\usepackage{multirow}
\usepackage{colortbl}
\usepackage{color}
\usepackage{threeparttable}
\usepackage{booktabs}
\usepackage{adjustbox}
\usepackage{subfig}
\usepackage{caption}
\usepackage[misc]{ifsym}
\usepackage{xcolor,colortbl}
\usepackage{makecell}
\usepackage{color}
\usepackage{tikz}
\usepackage[edges]{forest}
\definecolor{hidden-draw}{RGB}{20,68,106}
\definecolor{hidden-pink}{RGB}{255,245,247}
\usepackage{placeins}
\usepackage[linesnumbered,lined,boxed,commentsnumbered,ruled]{algorithm2e}

\begin{document}

\title{A Survey on Video Diffusion Models}


\author{Zhen Xing}
\email{zxing20@fudan.edu.cn}
\author{Qijun Feng}
\email{qjfeng21@m.fudan.edu.cn}
\author{Haoran Chen}
\email{chenhran21@m.fudan.edu.cn}
\affiliation{%
  \institution{Shanghai Collaborative Innovation Center of Intelligent Visual Computing and School of Computer Science, Fudan University}
  \country{China}
}

\author{Qi Dai}
\email{qid@microsoft.com}
\author{Han Hu}
\email{ancientmooner@gmail.com}
\affiliation{%
  \institution{Microsoft Research Asia}
  \country{China}}

\author{Hang Xu}
\affiliation{%
  \institution{Noah’s Ark Lab}
  \country{China}
}

\author{Zuxuan Wu}
\authornote{Corresponding author}
\email{zxwu@fudan.edu.cn}
\author{Yu-Gang Jiang}
\authornotemark[1]
\email{ygj@fudan.edu.cn}
\affiliation{%
  \institution{Shanghai Collaborative Innovation Center of Intelligent Visual Computing and School of Computer Science, Fudan University}
 \country{China}}

\renewcommand{\shortauthors}{Zhen Xing et al.}

\begin{abstract}
The recent wave of AI-generated content (AIGC) has witnessed substantial success in computer vision, with the diffusion model playing a crucial role in this achievement. Due to their impressive generative capabilities, diffusion models are gradually superseding methods based on GANs and auto-regressive Transformers, demonstrating exceptional performance not only in image generation and editing, but also in the realm of video-related research. However, existing surveys mainly focus on diffusion models in the context of image generation, with few up-to-date reviews on their application in the video domain.
To address this gap, this paper presents a comprehensive review of video diffusion models in the AIGC era. Specifically, we begin with a concise introduction to the fundamentals and evolution of diffusion models. Subsequently, we present an overview of research on diffusion models in the video domain, categorizing the work into three key areas: video generation, video editing, and other video understanding tasks. We conduct a thorough review of the literature in these three key areas, including further categorization and practical contributions in the field. Finally, we discuss the challenges faced by research in this domain and outline potential future developmental trends. A comprehensive list of video diffusion models studied in this survey is available at \url{https://github.com/ChenHsing/Awesome-Video-Diffusion-Models}.
\end{abstract}

\ccsdesc[500]{Computing methodologies~Computer vision}

\keywords{Survey, Video Diffusion Model, Video Generation, Video Editing, AIGC}

\received{18 Nov 2023}
\received[revised]{18 Jun 2024}
\received[accepted]{22 Aug 2024}

\maketitle

\vspace{-0.3cm}
\section{Introduction}
\begin{wrapfigure}[12]{r}{0.5\textwidth}
\centering
\vspace{-20pt}
\includegraphics[width=1.0\linewidth]{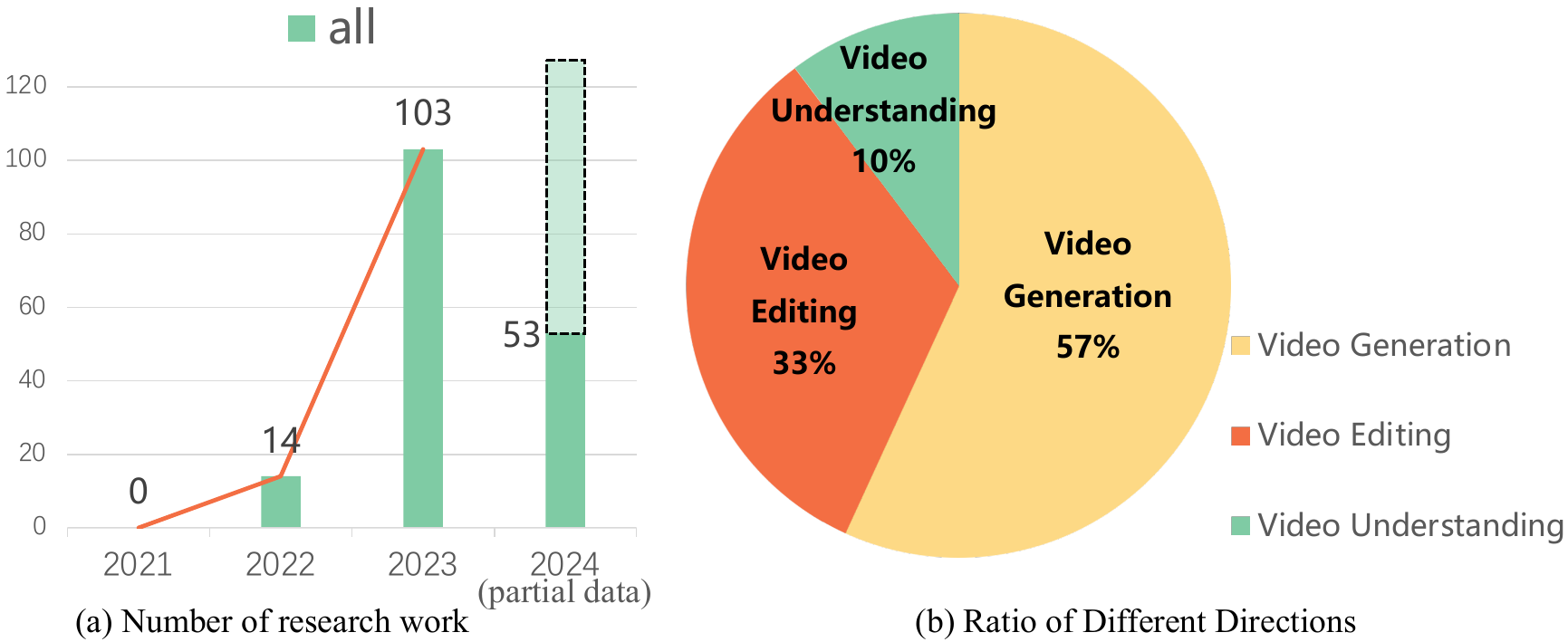}
\caption{Summarization on video diffusion model research works. (a) The number of related research works is rapidly increasing. (b) Video generation and editing are the top two research areas using diffusion models.}
\label{fig:number}
\end{wrapfigure}

AI-generated content (AIGC) is currently one of the most prominent research fields in computer vision and artificial intelligence. It has not only garnered extensive attention and scholarly investigation, but also exerted profound influence across industries and other applications, such as computer graphics, art and design, medical imaging, \emph{etc}. 
Among these endeavors, a series of approaches represented by diffusion models~\cite{stablediffusion,  dalle2, Midjourney, controlnet, ruiz2023dreambooth, cascaded, imagen} have emerged as particularly successful, rapidly supplanting methods based on generative adversarial networks (GANs)~\cite{gan,aliasgan,stylegan,stylegan2,stylegan3} and auto-regressive Transformers~\cite{taming,zeroshot,scaling, ding2022cogview2} to become the predominant approach for image generation.
Due to their strong controllability, photorealistic generation, and impressive diversity, diffusion-based methods 
also bloom across a wide range of computer vision tasks, including
image editing~\cite{brooks2023instructpix2pix, hertz2022prompt2prompt, meng2021sdedit, tumanyan2023plug}, dense prediction~\cite{diffusiondet, diffusioninst, diffmatch, ddp, dformer}, and diverse areas such as video synthesis~\cite{singer2022make, vdm, ho2022imagenvideo, videofusion, videofactory, SimDA} and 3D generation~\cite{magic3d, dreamfusion, diffusion3d, feng2024fdgaussian}.
As one of the most important mediums, video emerges as a dominant force on the Internet. Compared to mere text and static image, video presents a trove of dynamic information, providing users with a more comprehensive and immersive visual experience. Research on video tasks based on the diffusion models is progressively gaining traction. As shown in Fig.~\ref{fig:number}, the number of research publications of video diffusion models has increased significantly since 2022 and can be categorized into three major classes: video generation~\cite{singer2022make, vdm, videoLDM,SimDA,  Text2video-zero, videofusion, videofactory}, video editing~\cite{tuneavideo, gen1, qi2023fatezero, magicedit, ccedit}, and  video understanding~\cite{difftad, pix2seq-D, momentdiff, wang2023pdpp}.


With the rapid advancement of video diffusion models~\cite{vdm} and their demonstration of impressive results, the endeavor to track and compare recent research on this topic gains great importance. Several survey articles have covered foundational models in the era of AIGC~\cite{aigcSurvey, wu2023aigcSurvey}, encompassing the diffusion model itself~\cite{diffusionmodelsurvey, efficientdiffusionsurvey} and multi-modal learning~\cite{zhan2023multimodalsurvey, vlsurvey, vlsurvey2}. There are also surveys specifically focusing on text-to-image~\cite{t2isurvey} research and text-to-3D~\cite{text-to-3d} applications. However, these surveys either provide only a coarse coverage of the video diffusion models or place greater emphasis on image models~\cite{efficientdiffusionsurvey, zhan2023multimodalsurvey, t2isurvey}. 
As such, in this work, we aim to fulfill the blank with a comprehensive review of the methodologies, experimental settings, benchmark datasets, and other video applications of the diffusion model.

\noindent\textbf{Contribution}: In this survey, we systematically track and summarize recent literature concerning video diffusion models, encompassing domains such as video generation, editing, and other aspects of video understanding. By extracting shared technical details, this survey covers the most representative works in the field. Background and relevant knowledge preliminaries concerning video diffusion models are also introduced. Furthermore, we conduct a comprehensive analysis and comparison of benchmarks and settings for video generation. To the best of our knowledge, we are the first to concentrate on this specific domain. More importantly, given the rapid evolution of the video diffusion, we might not cover all the latest advancements in this survey. Therefore we encourage researchers to get in touch with us to share their new findings in this domain, enabling us to maintain currency. These novel contributions will be incorporated into the revised version for discussion.

\noindent\textbf{Survey Pipeline}: In Section~\ref{Sec:preliminaries}, we will cover background knowledge, including problem definition, datasets, evaluation metrics, and relevant research domains. Subsequently, in Section~\ref{Sec:videogeneration}, we primarily present an overview of methods in the field of video generation. In Section~\ref{Sec:videoediting}, we delve into the principal studies concerning video editing tasks. In Section~\ref{Sec:video-understanding}, we elucidate the various directions of utilizing diffusion models for video understanding. In Section~\ref{Sec:future}, we highlight the existing research challenges and potential future avenues, culminating in our concluding remarks in Section~\ref{Sec:conclusion}.

\section{Preliminaries}
\label{Sec:preliminaries}
In this section, we first present preliminaries of diffusion models, followed by reviewing the related research domains. Finally, we introduce the commonly used datasets and evaluation metrics.

\subsection{Diffusion Model}
\begin{figure*}[!t]
    \centering
    \includegraphics[width=0.88\textwidth]{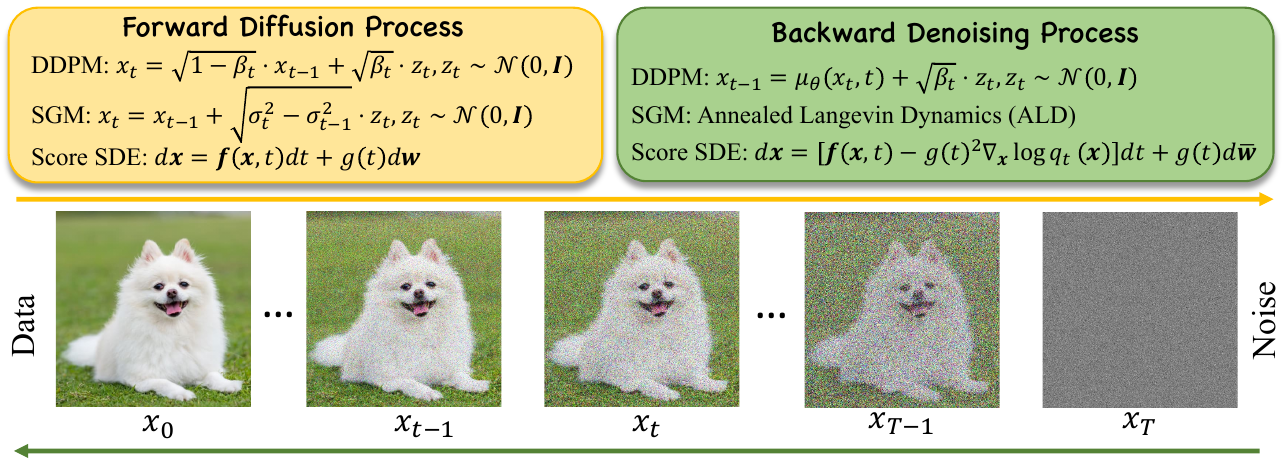}
    \vspace{-0.2cm}
    \caption{\textbf{Overview of diffusion model.} We demonstrate the diffusion and denoising process of DDPM, SGM and Score SDE.}
    \label{fig:diffusion}
\end{figure*}

Diffusion models~\cite{ddpm, song2020score} are a category of probabilistic generative models that learn to reverse a process that gradually degrades the training data structure and have become the new state-of-the-art family of deep generative models. They have broken the long-held dominance of generative adversarial networks (GANs)~\cite{GANs} in a variety of challenging tasks such as image generation~\cite{sohl2015deep, song2019generative, rombach2022high, song2020improved, nichol2021glide, vahdat2021score}, image super-resolution~\cite{rombach2022high, saharia2022image, batzolis2021conditional, kawar2022denoising}, and image editing~\cite{avrahami2022blended, choi2021ilvr}. Current research on diffusion models is mostly based on three predominant formulations: denoising diffusion probabilistic models (DDPMs)~\cite{ddpm, improvedddpm, sohl2015deep}, score-based generative models (SGMs)~\cite{song2019generative, song2020improved}, and stochastic differential equations (Score SDEs)~\cite{song2020score, song2021maximum}. \textcolor{black}{The diffusion and denoising processes of these three formulations are summarized and demonstrated in Fig.~\ref{fig:diffusion}}.

\subsubsection{Denoising Diffusion Probabilistic Models (DDPMs)}
\label{Sec:DDPMs}

A \textit{denoising diffusion probabilistic model} (DDPM)~\cite{ddpm, improvedddpm, sohl2015deep} involves two Markov chains: a forward chain that perturbs data to noise, and a reverse chain that converts noise back to data. The former aims at transforming any data into a simple prior distribution, while the latter learns transition kernels to reverse the former process. New data points can be generated by first sampling a random vector from the prior distribution, followed by ancestral sampling through the reverse Markov chain. The pivot of this sampling process is to train the reverse Markov chain to match the actual time reversal of the forward Markov chain.

Formally, given a data distribution $x_{0} \backsim q(x_{0})$, the forward Markov process generates a sequence of random variables $x_{1}, x_{2},...,x_{T}$ with transition kernel $q(x_{t}|x_{t-1})$. The joint distribution of $x_{1}, x_{2},...,x_{T}$ conditioned on $x_{0}$, denoted as $q(x_{1},...,x_{T}|x_{0})$, can be factorized into
\begin{equation}
    q(x_{1},...,x_{T}|x_{0})=\displaystyle\prod_{t = 1}^{T}q(x_t|x_{t-1}).
\end{equation}
Typically, the transition kernel is designed as
\begin{equation}
    q(x_t|x_{t-1}) = \mathcal{N}(x_t;\sqrt{1-\beta_t}x_{t-1}, \beta_t \textbf{I}),
\end{equation}
where $\beta_t \in (0,1)$ is a hyperparameter chosen ahead of model training.

The reverse Markov chain is parameterized by a prior distribution $p(x_T)=\mathcal{N}(x_T; 0, \textbf{I})$ and a learnable transition kernel $p_{\theta}(x_{t-1}|x_t)$ which takes the form of
\begin{equation}
    p_{\theta}(x_{t-1}|x_t)=\mathcal{N}(x_{t-1}; \mu_{\theta}(x_t, t), \Sigma_{\theta}(x_t, t))
\end{equation} 
where $\theta$ denotes model parameters and the mean $\mu_{\theta}(x_t, t)$ and variance $\Sigma_{\theta}(x_t, t)$ are parameterized by deep neural networks. 
With the reverse Markov chain, we can generate new data $x_0$ by first sampling a noise vector $x_T \backsim p(x_T)$, then iteratively sampling from the learnable transition kernel $x_{t-1} \backsim p_{\theta}(x_{t-1}|x_t)$ until $t = 1$.

\subsubsection{Score-Based Generative Models (SGMs)}
\label{Sec:SGMs}

The key idea of score-based generative models (SGMs)~\cite{song2019generative, song2020improved} is to perturb data using various levels of noise and simultaneously estimate the scores corresponding to all noise levels by training a single conditional score network. Samples are generated by chaining the score functions at decreasing noise levels with score-based sampling approaches. Training and sampling are entirely decoupled in the formulation of SGMs.

With similar notations in Sec.~\ref{Sec:DDPMs}, let $q(x_0)$ be the data distribution, and $0 < \sigma_1 < \sigma_2 < ... < \sigma_T$ be a sequence of noise levels. A typical example of SGMs involves perturbing a data point $x_0$ to $x_t$ by the Gaussian noise distribution $q(x_t|x_0)= \mathcal{N}(x_t;x_0, \sigma_{t}^{2}I)$, which yields a sequence of noisy data densities $q(x_1), q(x_2), ..., q(x_T)$, where $q(x_t) := \int q(x_t)q(x_0)dx_0$. A noise-conditional score network (NCSN) is a deep neural network $s_\theta(x, t)$ trained to estimate the score function $\nabla_{x_t}\log q(x_t)$. We can directly employ techniques such as score matching, denoising score matching, and sliced score matching to train our NCSN from perturbed data points.

For sample generation, SGMs leverage iterative approaches to produce samples from $s_{\theta}(x, T), s_{\theta}(x, T-1), ..., s_{\theta}(x, 0)$ in succession by using techniques such as annealed Langevin dynamics (ALD).

\subsubsection{Stochastic Differential Equations (Score SDEs)}
\label{Sec:SDEs}
Perturbing data with multiple noise scales is key to the success of the above methods. Score SDEs~\cite{song2020score} generalize this idea further to an infinite number of noise scales. The diffusion process can be modeled as the solution to the following stochastic differential equation (SDE):
\begin{equation}
    d\textbf{x} = \textbf{f}(\textbf{x}, t)dt + g(t)d\textbf{w}
\end{equation}
where $\textbf{f}(\textbf{x}, t)$ and $g(t)$ are diffusion and drift functions of the SDE, and $\textbf{w}$ is a standard Wiener process. 

Starting from samples of $\textbf{x}(T) \backsim p_T$ and reversing the process, we can obtain samples $\textbf{x}(0) \backsim p_0$ through this reverse-time SDE:
\begin{equation}\label{reverse-SDE}
    d\textbf{x} = [\textbf{f}(\textbf{x}, t) - g(t)^{2}\nabla_{\textbf{x}} \log q_t(\textbf{x})]dt + g(t)d\bar{\textbf{w}}
\end{equation}
where $\bar{\textbf{w}}$ is a standard Wiener process when time flows backwards. Once the score of each marginal distribution, $\nabla_{\textbf{x}} \log p_t(\textbf{x})$, is known for all $t$, we can derive the reverse diffusion process from Eq.(\ref{reverse-SDE}) and simulate it to sample from $p_0$.

\subsection{Related Tasks}

The applications of video diffusion model contain a wide scope of video tasks, including video generation, video editing, and various other forms of video understanding. The methodologies for these tasks share similarities, often formulating the problems as diffusion generation tasks or utilizing the potent controlled generation capabilities of diffusion models for downstream tasks. In this survey, the main focus lies on the tasks such as Text-to-Video generation~\cite{singer2022make, ho2022imagenvideo, SimDA}, unconditional video generation~\cite{vdt, lamd, VIDM}, and text-guided video editing~\cite{gen1, tuneavideo, dreamix}, \emph{etc.}

\noindent$\bullet$ \textbf{Text-to-Video Generation} aims to automatically generate corresponding videos based on the textual descriptions. This typically involves comprehending the scenes, objects, and actions within the textual descriptions and translating them into a sequence of coherent visual frames, resulting in a video with both logical and visual consistency. T2V has broad applications, including the automatic generation of movies~\cite{zhu2023moviefactory}, animations~\cite{Animate-A-Story, AnimateDiff}, virtual reality content, educational demonstration videos~\cite{videoAdapter}, \emph{etc}.

\noindent$\bullet$ \textbf{Unconditional Video Generation} is a generative task where the objective is to generate a continuous and visually coherent sequence of videos starting from random noise or a fixed initial state, without relying on specific input conditions. Unlike conditional video generation, unconditional video generation does not require any external guidance or prior information~\cite{videofusion,lvdm,vdm}. The generative model needs to learn how to capture temporal dynamics, actions, and visual coherence in the absence of explicit inputs, to produce video content that is both realistic and diverse. This is crucial for exploring the ability of generative models to learn video content from unsupervised data and showcase diversity. 

\noindent$\bullet$ \textbf{Text-guided Video Editing} is a technique that involves using textual descriptions to guide the process of editing video content. In this task, a natural language description is provided as input, describing the desired changes or modifications to be applied to a video. The system then analyzes the textual input, extracts relevant information such as objects, actions, or scenes, and uses this information to guide the editing process.
Text-guided video editing offers a way to facilitate efficient and intuitive editing by allowing editors to communicate their intentions using natural language~\cite{gen1, videofusion,shin2023editavideo}, potentially reducing the need for manual and time-consuming frame-by-frame editing.

\subsection{Datasets and Metrics}
\label{Sec:data-evaluation}

\subsubsection{Data}
\label{Sec:data}

The evolution of video understanding tasks often aligns with the development of video datasets, and the same applies to video generation tasks. In the early stages of video generation, tasks are limited to training on low-resolution~\cite{moving-mnist}, small-scale datasets to specific domains~\cite{taichi-hd, time-lapse}, resulting in relatively monotonous video generation. With the emergence of large-scale video-text paired datasets, tasks such as general text-to-video generation~\cite{vdm, singer2022make} began to gain traction. Thus, the datasets of video generation can be mainly categorized into caption-level and category-level, as will be discussed separately.

\begin{table}[]
    \centering
    \scalebox{0.7}{

\begin{tabular}{@{}rrrrrr@{}}
\toprule
Dataset          &Year        & Text      & Domain   & \#Clips & Resolution \\ \midrule
MSR-VTT~\cite{msrvtt} & 2016        & Manual    & Open     & 10K           & 240P       \\
DideMo~\cite{DideMo}  & 2017        & Manual    & Flickr   & 27K           & -          \\
LSMDC~\cite{LSMDC}  & 2017         & Manual    & Movie    & 118K          & 1080P      \\
ActivityNet~\cite{ANet-caption} & 2017   & Manual    & Action   & 100K          & -          \\
YouCook2~\cite{YouCook2} & 2018       & Manual    & Cooking  & 14K           & -          \\
How2~\cite{How2}      & 2018      & Manual    & Instruct & 80K           & -          \\
VATEX~\cite{vatex}   &2019 &Manual &Action & 41K & 240P \\
HowTo100M~\cite{howto100m} & 2019      & ASR       & Instruct & 136M          & 240P       \\
WTS70M~\cite{WTS70M}  &  2020       & Metadata  & Action   & 70M           & -          \\
YT-Temporal~\cite{YT-temporal-180M} &2021   & ASR       & Open     & 180M          & -          \\
WebVid10M~\cite{webvid}  & 2021     & Alt-text  & Open     & 10.7M         & 360P       \\
Echo-Dynamic~\cite{scm} & 2021  & Manual  & Echocardiogram & 10K  & - \\
Tiktok~\cite{DisCo}   & 2021  & Mannual   & Action        & 0.3K     & -        \\
HD-VILA~\cite{hd-vila} & 2022   & ASR       & Open     & 103M          & 720P       \\
VideoCC3M~\cite{VideoCC3M}   & 2022    & Transfer  & Open     & 10.3M         & -          \\
HD-VG-130M~\cite{videofactory} & 2023     & Generated & Open     & 130M          & 720P       \\
InternVid~\cite{wang2023internvid} & 2023      & Generated & Open     & 234M          & 720P    \\
CelebV-Text~\cite{yu2023celebv} & 2023      & Generated & Face     & 70K          & 480P    \\
Panda-70M~\cite{chen2024panda} & 2024      & Generated & Open     & 70.8M          & 720P  
\\ \bottomrule
\end{tabular}
\label{Tab:video-text-datasets}

}
    \caption{The comparison of main caption-level video datasets.}
    \vspace{-1cm}
    \label{Tab:video-text-datasets}
\end{table}

\begin{table}[]
    \centering
\scalebox{0.7}{
\begin{tabular}{@{}rrrrrl@{}}
\toprule
Datasets       & Year & Categories & \#Clips  & Resolution                                                                \\ \midrule
UCF-101~\cite{ucf101}        & 2012 & 101        & 13K   &  $256\times256$   \\
Cityscapes~\cite{cordts2016cityscapes}     & 2015 & 30         & 3K    & $256\times256$                                                    \\
Moving MNIST~\cite{moving-mnist}   & 2016 & 10         & 10K  & $64\times64$                           \\
Kinetics-400~\cite{kinetic}   & 2017 & 400        & 260K & $256\times256$   \\
BAIR~\cite{BAIR}           & 2017 & 2          & 45K  & $64\times64$                                \\
DAVIS~\cite{davis}           & 2017 & -          & 90  & $1280\times720$ \\ 
Sky Time-Lapse~\cite{time-lapse} & 2018 & 1          & 38K    & $256\times256$     \\
Ssthv2~\cite{goyal2017somethingv2}         & 2018 & 174        & 220K   & $256\times256$                            \\
Kinetics-600~\cite{kinetics600}   & 2018 & 600        & 495K  & $256\times256$    \\
Epic-Kitchen~\cite{epickitchen}   & 2018 & 149        & 90K  & $1920\times1080$  \\
Tai-Chi-HD~\cite{taichi-hd}     & 2019 & 1          & 3K   & $256\times256$                      \\
Bridge Data~\cite{bridgedata}         & 2021 & 10         & 7K   & $256\times256$                                                 \\ 
Mountain Bike~\cite{mountainbike}  & 2022 & 1          & 1K   & $576\times1024$                         \\
RDS~\cite{videoLDM}            & 2023 & 2          & 683K  & $512\times1024$               \\ \bottomrule
\end{tabular}
}
\label{Tab:category-level}
    \caption{The comparison of existing category-level datasets for video generation and editing.}
    \vspace{-1cm}
    \label{Tab:category-level}
\end{table}


\noindent$\bullet$ \textbf{Caption-level Datasets} consist of videos paired with descriptive text captions, providing essential data for training models to generate videos based on textual descriptions.
We list several common caption-level datasets in Table~\ref{Tab:video-text-datasets}, which vary in scale and domain. Early caption-level video datasets were primarily used for video-text retrieval tasks~\cite{msrvtt,DideMo, LSMDC}, with small-scales (less than 120K) and a limited focus on specific domains (\emph{e.g.} movie~\cite{LSMDC}, action~\cite{ANet-caption, WTS70M}, cooking~\cite{YouCook2}). With the introduction of the open-domain WebVid-10M~\cite{webvid} dataset, a new task of text-to-video (T2V) generation gains momentum, leading researchers to focus on open-domain T2V generation tasks.
Despite being a mainstream benchmark dataset for T2V tasks, it still suffers from issues such as low resolution (360P) and watermarked content.
Subsequently, to enhance the resolution and broader coverage of videos in the general text-to-video (T2V) tasks, Panda-70M~\cite{chen2024panda}, VideoFactory~\cite{videofactory} and InternVid~\cite{wang2023internvid} introduce larger-scale (70M \& 130M \& 234M) and high-definition (720P) open-domain datasets. To collect diverse video datasets, 
VidRD~\cite{VidRD} utilizes static images~\cite{laion400m}, long videos~\cite{videolt} and short videos~\cite{kinetic} when constructing the video-text dataset.

\noindent$\bullet$ \textbf{Category-level Datasets} consist of videos grouped into specific categories, with each video labeled by its category. The datasets are commonly utilized for unconditional video generation or class conditional video generation tasks.
We summarize category-level commonly used video datasets in Table~\ref{Tab:category-level}. Notably, several of these datasets 
are also applied to other tasks. For instance, UCF-101~\cite{ucf101}, Kinetics~\cite{kinetic,kinetics600}, and Something-Something~\cite{goyal2017somethingv2} are typical benchmarks for action recognition. DAVIS~\cite{davis} was initially proposed for the video object segmentation task and later became a commonly used benchmark for video editing.
Among these datasets, UCF-101~\cite{ucf101} stands out as the most widely utilized in video generation, serving as a benchmark for unconditional video generation, category-based conditional generation, and video prediction applications. It comprises samples from YouTube~\cite{youtube} that encompasses 101 action categories, including human sports, musical instrument playing, and interactive actions. 
Akin to UCF, Kinetics-400~\cite{kinetic} and Kinetics-600~\cite{kinetics600} are two datasets 
encompassing more complex action categories and larger data scale, while retaining the same application scope as UCF-101~\cite{ucf101}. The Something-Something~\cite{goyal2017somethingv2} dataset, on the other hand, possesses both category-level and caption-level labels, rendering it particularly suitable for text-conditional video prediction tasks~\cite{Seer}. It is noteworthy that these sizable datasets that originally played pivotal roles in the realm of action recognition exhibit smaller scales (less than 50K) and single-category~\cite{time-lapse,taichi-hd}, single-domain attributes (digital number~\cite{moving-mnist}, driving scenery~\cite{cordts2016cityscapes, mountainbike, videoLDM}, egocentric~\cite{epickitchen}, robot~\cite{bridgedata}) and is thereby inadequate for producing high-quality videos. Consequently, in recent years, datasets specifically crafted for video generation tasks are proposed, typically originating from featuring unique attributes, such as high resolution (1080P)~\cite{videoLDM} or extended duration~\cite{mountainbike, nuwaxl}. For example, Long Video GAN~\cite{mountainbike} proposes dataset which has 66 videos with an average duration of 6504 frames at 30fps. Video LDM~\cite{videoLDM} collects RDS dataset consisting of 683,060 real driving videos of 8 seconds in length each with 1080P resolution.

\subsubsection{Evaluation Metrics}
\label{Sec:Evaluation}
Evaluation metrics for video generation are commonly categorized into quantitative and qualitative measures. For qualitative measures, human subjective evaluation has been used in several works~\cite{singer2022make, videofactory,tuneavideo, SimDA}, where evaluators are typically presented with two or more generated videos to compare against videos synthesized by other competitive models. Observers generally engage in voting-based assessments regarding the realism, natural coherence, and text alignment of the videos (T2V tasks). However, human evaluation is both costly and at the risk of failing to reflect the full capabilities of the model~\cite{alqahtani2019analysis}. Therefore, in the following we will primarily delve into the quantitative evaluation standards for image-level and video-level assessments. 

\noindent$\bullet$ \textbf{Image-level Metrics}
Videos are composed of a sequence of image frames, thus image-level evaluation metrics can provide a certain amount of insight into the quality of the generated video frames. Commonly employed image-level metrics include Fréchet Inception Distance (FID)~\cite{fid}, Peak Signal-to-Noise Ratio (PSNR)~\cite{SSIM}, Structural Similarity Index (SSIM)~\cite{SSIM}, and CLIPSIM~\cite{clip}. FID~\cite{fid} assesses the quality of generated videos by comparing synthesized video frames to real video frames. It involves preprocessing the images for normalization to a consistent scale, utilizing InceptionV3~\cite{incpetionv3} to extract features from real and synthesized videos, and computing mean and covariance matrices. These statistics are then combined to calculate the FID~\cite{fid} score.

Both SSIM~\cite{SSIM} and PSNR~\cite{SSIM} are pixel-level metrics. SSIM~\cite{SSIM} evaluates brightness, contrast, and structural features of original and generated images, while PSNR~\cite{SSIM} is a coefficient representing the ratio between peak signal and Mean Squared Error (MSE)~\cite{mse}. These two metrics are commonly used to assess the quality of reconstructed image frames, and are applied in tasks such as super-resolution and in-painting.
CLIPSIM~\cite{clip} is a method for measuring image-text relevance. 
Based on the CLIP~\cite{clip} model, it extracts both image and text features and then computes the similarity between them.
This metric is often employed in text-conditional video generation or editing tasks~\cite{singer2022make, SimDA, videofactory, videoLDM, tuneavideo, save}.

\noindent$\bullet$ \textbf{Video-level Metrics}
Although image-level evaluation metrics represent the quality of generated video frames, they primarily focus on individual frames, disregarding the temporal coherence of the video. Video-level metrics, on the other hand, would provide a more comprehensive evaluation of video generation. Fréchet Video Distance (FVD)~\cite{fvd} is a video quality evaluation metric based on FID~\cite{fid}. Unlike image-level methods that use the Inception~\cite{incpetionv3} network to extract features from a single frame, FVD~\cite{fvd} employs the Inflated-3D Convnets (I3D)~\cite{kinetic} pre-trained on Kinetics~\cite{kinetic} to extract features from video clips. Subsequently, FVD scores are computed through the combination of means and covariance matrices. Kernel Video Distance (KVD)~\cite{kvd} is also based on I3D~\cite{kinetic} features, but it differentiates itself by utilizing Maximum Mean Discrepancy (MMD)~\cite{MMD}, a kernel-based method, to assess the quality of generated videos. Video IS (Inception Score)~\cite{videoIS} calculates the Inception score of generated videos using features extracted by the 3D-Convnets (C3D)~\cite{c3d}, which is often applied in evaluation on UCF-101~\cite{ucf101}. High-quality videos are characterized by a low entropy probability, denoted as $P(y|x)$, whereas diversity is assessed by examining the marginal distribution across all videos, which should exhibit a high level of entropy.
Frame Consistency CLIP Score~\cite{clip} is often used in video editing tasks~\cite{tuneavideo, controlvideo, SimDA} to measure the coherence of edited videos. It is calculated by obtaining CLIP image embeddings for all frames and averaging the cosine similarity between all pairs of frames.

\section{Video Generation}
\label{Sec:videogeneration}

\tikzstyle{my-box}=[
    rectangle,
    draw=hidden-draw,
    rounded corners,
    text opacity=1,
    minimum height=1.5em,
    minimum width=5em,
    inner sep=2pt,
    align=center,
    fill opacity=.5,
    line width=0.8pt,
]
\tikzstyle{leaf}=[my-box, minimum height=1.5em,
    fill=hidden-pink!80, text=black, align=left,font=\tiny,
    inner xsep=2pt,
    inner ysep=4pt,
    line width=0.8pt,
]
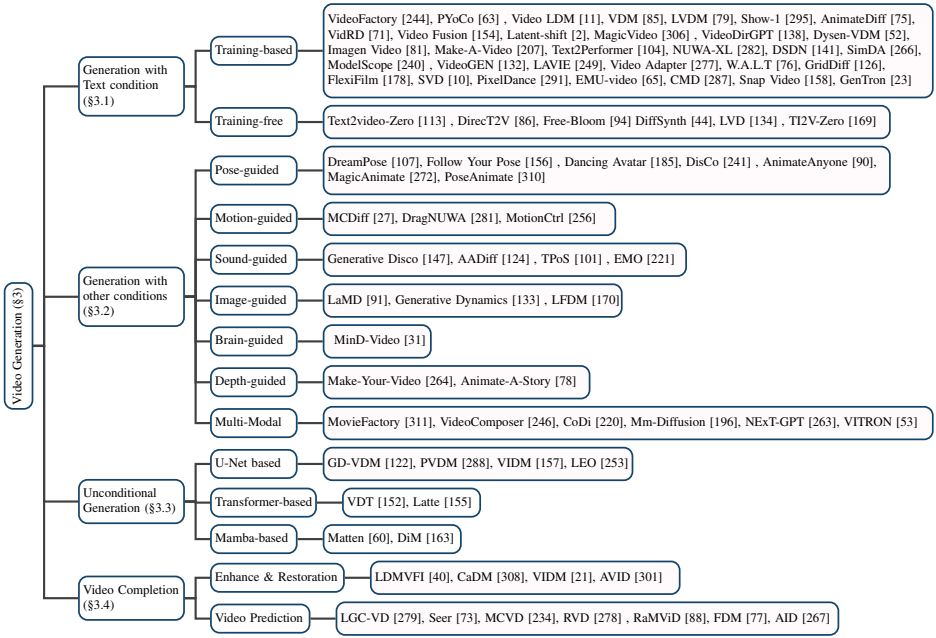
\begin{figure*}[t!]
    \centering
    \resizebox{0.9\textwidth}{!}{
        \begin{forest}
            forked edges,
            for tree={
                grow=east,
                reversed=true,
                anchor=base west,
                parent anchor=east,
                child anchor=west,
                base=left,
                font=\tiny,
                rectangle,
                draw=hidden-draw,
                rounded corners,
                align=left,
                minimum width=2em,
                edge+={darkgray, line width=1pt},
                s sep=3pt,
                inner xsep=2pt,
                inner ysep=3pt,
                line width=0.8pt,
                ver/.style={rotate=90, child anchor=north, parent anchor=south, anchor=center},
            },
            where level=1{text width=4.5em,font=\tiny,}{},
            where level=2{text width=3.6em,font=\tiny,}{},
            where level=3{text width=3.0em,font=\tiny,}{},
            [
                Video Generation (\S \ref{Sec:videogeneration}), ver
                [
                    Generation with \\ Text condition \\ (\S \ref{Sec:generalT2V})
                    [
                        Training-based
                        [
                           VideoFactory~\cite{videofactory}{,}  PYoCo~\cite{pyoco} {,} Video LDM~\cite{videoLDM}{,} VDM~\cite{vdm}{,} LVDM~\cite{lvdm}{,}  Show-1~\cite{show1}{,} AnimateDiff~\cite{AnimateDiff}{,}   \\
                           VidRD~\cite{VidRD}{,}
                           Video Fusion~\cite{videofusion}{,} 
                            Latent-shift~\cite{latentshift}{,} MagicVideo~\cite{magicvideo} {,}  VideoDirGPT~\cite{videodirectorgpt}{,}   Dysen-VDM~\cite{Dysen-VDM}{,}
                            \\ 
                             Imagen Video~\cite{ho2022imagenvideo}{,}  Make-A-Video~\cite{singer2022make}{,} Text2Performer~\cite{Text2Performer}{,}    NUWA-XL~\cite{nuwaxl}{,}  DSDN~\cite{DSDN}{,} SimDA~\cite{SimDA}{,}
                             \\  
                            ModelScope~\cite{modelscope} {,}  VideoGEN~\cite{videogen}{,} 
                             LAVIE~\cite{lavie}{,} Video Adapter~\cite{videoAdapter}{,} \textcolor{black}{W.A.L.T~\cite{walt}{,}  GridDiff~\cite{griddiff}}{,} \\  \textcolor{black}{FlexiFilm~\cite{FlexiFilm}{,} SVD~\cite{svd}{,} PixelDance~\cite{pixeldance}{,} EMU-video~\cite{emuvideo}{,} CMD~\cite{cmd}{,} Snap Video~\cite{snapvideo}{,} GenTron~\cite{gentron}
                             }
                            , leaf, text width=28em
                        ]
                    ]
                    [
                        Training-free
                        [
                            Text2video-Zero~\cite{Text2video-zero} {,} DirecT2V~\cite{DirecT2V}{,}
                            Free-Bloom~\cite{freebloom} 
                            DiffSynth~\cite{diffsynth}{,} LVD~
                            \cite{LVD} {,} 
                            \textcolor{black}{TI2V-Zero~\cite{ti2v-zero}}
                            , leaf, text width=26em
                        ]
                    ]
                ]
                [
                    Generation with \\ other conditions                  
                    \\ (\S \ref{Sec:modality-control})
                    [
                        Pose-guided
                        [
                            DreamPose~\cite{Dreampose}{,} Follow Your Pose~\cite{FollowYourPose} {,} Dancing Avatar~\cite{DancingAvatar}{,} DisCo~\cite{DisCo} {,} \textcolor{black}{AnimateAnyone~\cite{hu2024animateanyone}}{,} \\ \textcolor{black}{MagicAnimate~\cite{xu2024magicanimate}{,} PoseAnimate~\cite{poseanimate}}
                            , leaf, text width=26em
                        ]
                    ]
                    [
                        Motion-guided
                        [
                             MCDiff~\cite{MCDiff}{,} DragNUWA~\cite{DragNUWAFC}{,} \textcolor{black}{MotionCtrl~\cite{motionctrl}}
                            , leaf, text width=13.2em
                        ]
                    ]
                    [
                        Sound-guided
                        [
                            Generative Disco~\cite{generativeDisco}{,} AADiff~\cite{aadiff} {,}
                            TPoS~\cite{TPoS} {,} \textcolor{black}{EMO~\cite{emo}}
                            , leaf, text width=16.5em
                        ]
                    ]
                    [
                        Image-guided
                        [
                            LaMD~\cite{lamd}{,} Generative Dynamics~\cite{generativeDynamics}
                            {,} LFDM~\cite{LFDM} \\
                            , leaf, text width=13.5em
                        ]
                    ]
                    [
                        Brain-guided
                        [
                            MinD-Video~\cite{Cinematic}
                            , leaf, text width=4em
                        ]
                    ]
                    [
                        Depth-guided
                        [
                            Make-Your-Video~\cite{Make-Your-Video}{,} Animate-A-Story~\cite{Animate-A-Story}
                            , leaf, text width=12em
                        ]
                    ]
                    [
                        Multi-Modal 
                        [
                            MovieFactory~\cite{zhu2023moviefactory}{,}
                            VideoComposer~\cite{videocomposer}{,} CoDi~\cite{CoDi}{,} Mm-Diffusion~\cite{mm-diffusion}{,} NExT-GPT~\cite{next-gpt}{,} \textcolor{black}{VITRON~\cite{vitron}}                            , leaf, text width=28em
                        ]
                    ]
                ]
                [
                    Unconditional 
                    \\Generation (\S \ref{Sec:unconditional})
                    [
                        U-Net based
                        [
                            GD-VDM~\cite{gd-vdm}{,} PVDM~\cite{PVDM}{,} VIDM~\cite{VIDM}{,} LEO~\cite{LEO}  \\ 
                            , leaf, text width=14em
                        ]
                    ]
                    [
                        Transformer-based , text width= 4.5em
                        [
                            VDT~\cite{vdt}{,} \textcolor{black}{Latte~\cite{latte}}
                            ,  text width=6em
                        ]
                    ]
                    [
                        Mamba-based
                        [
                           \textcolor{black}{ Matten~\cite{matten}{,} DiM~\cite{DiM}}
                            ,  text width=6em
                        ]
                    ]
                ]
                [
                    Video Completion
                    \\ (\S \ref{Sec:video-completion})
                    [
                        Enhance \& Restoration, text width=5.8em
                        [
                            LDMVFI~\cite{LDMVFI}{,} CaDM~\cite{CaDM}{,} VIDM~\cite{lookvidm}{,} \textcolor{black}{AVID~\cite{avid}}
                            , leaf, text width=14em
                        ]
                    ]
                    [
                        Video Prediction, text width=4.2em
                        [
                            LGC-VD~\cite{LGCVD}{,} Seer~\cite{Seer}{,} MCVD~\cite{mcvd}{,} RVD~\cite{RVD} {,} RaMViD~\cite{RaMViD}{,} FDM~\cite{FDM}{,} \textcolor{black}{AID~\cite{aid}}
                            , leaf, text width=23.0em
                        ]
                    ]
                ]
            ]
        \end{forest}
    }
    \vspace{-0.3cm}
    \caption{Taxonomy of Video Generation. Key aspects of Video Generation include General T2V Generation, Domain-specific Generation, Conditional Control Generation, and Video Completion. }
    \label{taxo_of_videogeneration}
    \vspace{-0.6cm}
\end{figure*}

In this section, we categorize video generation into four groups and provide detailed reviews for each:  Text-to-video~(T2V) generation (Sec.~\ref{Sec:generalT2V}),  Video Generation with other conditions (Sec.~\ref{Sec:modality-control}), Unconditional Video Generation(Sec.~\ref{Sec:unconditional}) and Video Completion (Sec.~\ref{Sec:video-completion}). Finally, we summarize the settings and evaluation metrics, and present a comprehensive comparison of various models in Sec.~\ref{Sec:Benchmark}. The taxonomy details of video generation are demonstrated in Fig.~\ref{taxo_of_videogeneration}.

\subsection{Video Generation with Text Condition}

\label{Sec:generalT2V}
Evidenced by recent research~\cite{chatgpt, dalle2, stablediffusion}, the interaction between generative AI and natural language is of paramount importance. While significant progress has been achieved in generating images from text~\cite{dalle2, stablediffusion, Midjourney, ding2022cogview2}, the development of Text-to-Video (T2V) approaches is still in its early stages. In this section, we will introduce training-based and training-free T2V methods, respectively.



\subsubsection{Training-based T2V Diffusion Methods}
In the preceding discussion, we have briefly recapitulated a few T2V methods that do not rely on the diffusion model. Moving forward, we predominantly introduce the utilization of the currently most prominent diffusion model in the realm of T2V tasks.

\noindent$\bullet$ \textbf{Early T2V Exploration} Among the multitude of endeavors, VDM~\cite{vdm} stands as the pioneer in devising a video diffusion model for video generation. It extends the conventional image diffusion U-Net~\cite{unet} architecture to a 3D U-Net structure and employs joint training with both images and videos. The conditional sampling technique it employs enables generating videos of enhanced quality and extended duration. Being the first exploration of a diffusion model for T2V, it also accommodates tasks such as unconditional generation and video prediction. 

In contrast to VDM~\cite{vdm}, which requires paired \emph{video-text} datasets, Make-A-Video~\cite{singer2022make} introduces a novel paradigm. Here, the network learns visual-textual correlations from paired \emph{image-text} data and captures video motion from unsupervised video data. This innovative approach reduces the reliance on data collection, resulting in the generation of diverse and realistic videos. Furthermore, by employing multiple super-resolution models and interpolation networks, it achieves higher-definition and frame-rate generated videos. 

\noindent$\bullet$ \textbf{Temporal Modeling Exploration }
While previous approaches leverage diffusion at pixel-level, MagicVideo~\cite{magicvideo} stands as one of the earliest works to employ the Latent Diffusion Model (LDM)~\cite{stablediffusion} for T2V generation in latent space. By utilizing diffusion models in a lower-dimensional latent space, it significantly reduces computational complexity, thereby accelerating processing speed. The introduced frame-wise lightweight adaptor aligns the distributions of images and videos so that the proposed directed attention can better model temporal relationships to ensure video consistency. 

Concurrently, LVDM~\cite{lvdm} also employs the LDM~\cite{stablediffusion} as its backbone, utilizing a hierarchical framework to model the latent space. By employing a mask sampling technique, the model becomes capable of generating longer videos. It incorporates techniques such as Conditional Latent Perturbation~\cite{cascaded} and Unconditional Guidance~\cite{classifier-free} to mitigate performance degradation in the later stages of auto-regressive generation tasks. With this training approach, it can be applied to video prediction tasks, even generating long videos consisting of thousands of frames.

VideoFactory~\cite{videofactory} introduces a swapped cross-attention mechanism to facilitate interaction between the temporal and spatial modules, resulting in improved temporal relationship modeling. Besides, trained on its proposed HD-VG-130M dataset, the approach presented in the paper is capable of generating high-resolution videos at ($1376\times768$) resolution.

ModelScope~\cite{modelscope} incorporates spatial-temporal convolution and attention into LDM~\cite{stablediffusion} for T2V tasks. It adopts a mixed training approach using LAION~\cite{laion400m} and WebVid~\cite{webvid}, and serves as an open-source baseline method.

Previous methods predominantly rely on 1D convolutions or temporal attention~\cite{magicvideo} to establish temporal relationships. Latent-Shift~\cite{latentshift}, on the other hand, focuses on lightweight temporal modeling. Drawing inspiration from TSM~\cite{tsm}, it shifts channels between adjacent frames in convolution blocks for temporal modeling.  Additionally, the model maintains the original T2I~\cite{stablediffusion} capability while generating videos.

\noindent$\bullet$ \textbf{Multi-stage T2V methods}
Imagen Video~\cite{ho2022imagenvideo} extends the T2I model, Imagen~\cite{imagen}, for video generation using a cascaded video diffusion model composed of seven sub-models: one for base video generation, three for spatial super-resolution, and three for temporal super-resolution. This three-stage training pipeline employs T2I techniques like classifier-free guidance~\cite{classifier-free}, conditioning augmentation~\cite{cascaded}, and v-parameterization~\cite{v-para}. Progressive distillation techniques~\cite{v-para,progressuvedistillation} are used to speed up sampling time, making these multi-stage training techniques effective for high-definition video generation.

Concurrently, Video LDM~\cite{videoLDM} trains a T2V network composed of three training stages, including key-frame T2V generation, video frame interpolation and spatial super-resolution modules. It adds temporal attention layer and 3D convolution layer to the spatial layer, enabling the generation of key frames in the first stage. Subsequently, through the implementation of a mask sampling method, a frame interpolation model is trained, extending key frames of short videos to higher frame rates. Lastly, a video super-resolution model is employed to enhance the resolution. 

Similarly, LAVIE~\cite{lavie} employs a cascaded video diffusion model composed of three stages: a base T2V stage, a temporal interpolation stage, and a video super-resolution stage.
Furthermore, it validates that the process of joint image-video fine-tuning can yield high-quality and creative outcomes. 

Show-1~\cite{show1} introduces the fusion of pixel-based~\cite{deepfloyd} and latent-based~\cite{stablediffusion} diffusion models for T2V generation. Its framework has four stages: key frame generation, frame interpolation, and super-resolution at a low-resolution pixel level, followed by a latent super-resolution module to enhance video resolution cost-effectively. Pixel-level stages ensure precise text alignment in the videos. Latent-level stage offers a cost-effective means of enhancing video resolution.

\noindent$\bullet$ \textbf{Noise Prior Exploration}
While most of the methods mentioned denoising each frame independently through diffusion models, VideoFusion~\cite{videofusion} stands out by considering the content redundancy and temporal correlations among different frames.
Specifically, it decomposes the diffusion process using a shared base noise for each frame and residual noise along the temporal axis. This noise decomposition is achieved through two co-training networks. Such an approach is introduced to ensure consistency in generating frame motion, although it may lead to limited diversity. Furthermore, the paper shows that employing T2I backbones~\cite{dalle2} for training T2V models accelerates convergence, but its text embedding might face challenges in understanding long temporal sequences of text.

PYoCo~\cite{pyoco} acknowledges that directly extending the image noise prior to video can yield suboptimal outcomes in T2V tasks. As a solution, it intricately devises a video noise prior and fine-tune the eDiff-I~\cite{e-diff} model for video generation. The proposed noise prior involves sampling correlated noise for different frames within the video. The authors validate that the proposed mixed and progressive noise models are better suited for T2V tasks. 

FlexiFilm~\cite{FlexiFilm} introduces a temporal conditioner and a resampling strategy to maintain consistency between multimodal conditions and the generated videos. These strategies enhance consistency between generated videos and multimodal conditions, and address overexposure to produce videos up to 30 seconds long.

VidRD~\cite{VidRD} introduces the Reuse and Diffuse framework, which iteratively generates additional frames by reusing the original latent representations and following the previous diffusion process. In this way, it can generate long videos.

\noindent$\bullet$ \textbf{Efficient Training}
ED-T2V~\cite{edt2v} utilizes LDM~\cite{stablediffusion} as its backbone and freezes a substantial portion of parameters to reduce training costs. It introduces identity attention and temporal cross-attention to ensure temporal coherence. The approach proposed in this paper manages to lower training costs while maintaining comparable T2V generation performance.

SimDA~\cite{SimDA} devises a parameter-efficient training approach for T2V tasks by maintaining the parameter of T2I model~\cite{stablediffusion} fixed. It incorporates a lightweight spatial adapter for transferring visual information for T2V learning. Additionally, it introduces a temporal adapter to model temporal relationships in lower  feature dimensions. The proposed latent shift attention aids in maintaining video consistency. Moreover, the lightweight architecture enables speed up inference and makes it adaptable for video editing tasks.

To reduce the substantial computational resource requirements of 3D U-Nets, GridDiff~\cite{griddiff} proposes a method for video generation using a 2D U-Net. Specifically, this approach treats video as a grid image. This method allows for the straightforward extension of image generation and editing techniques to the video domain.

\noindent$\bullet$ \textbf{Personalized Video Generation}
Personalized video generation generally refers to creating videos tailored to a specific protagonist or style, addressing the generation of videos customized for personal preferences or characteristics. 
AnimateDiff~\cite{AnimateDiff} notices the success of LoRA~\cite{hu2021lora} and Dreambooth~\cite{ruiz2023dreambooth} in personalized T2I models and aims to extend their effectiveness to video animation. Furthermore, the authors aim at training a model that can be adapted to generate diverse personalized videos, without the need of repeatedly retraining on video datasets. This involves using a T2I model as a base generator and adding a motion module to learn motion dynamics. During inference, the personalized T2I model can replace the base T2I weights, enabling personalized video generation.

\noindent$\bullet$ \textbf{Image-conditioned T2V methods}
To address the issue of flickers and artifacts in T2V-generated videos, DSDN~\cite{DSDN} introduces a dual-stream diffusion model, one for video content and the other for motion. In this way, it can maintain a strong alignment between content and motion. By decomposing the video generation process into content and motion components, it is possible to generate continuous videos with fewer flickers.

VideoGen~\cite{videogen} first utilizes a T2I model~\cite{stablediffusion} to generate images based on the text prompt, which serves as a reference image for guiding video generation. Subsequently, an efficient cascaded latent diffusion module is introduced, employing flow-based temporal upsampling steps to enhance temporal resolution. 

Additionally, MicroCinema~\cite{microcinema}, PixelDance~\cite{pixeldance}, and EMU-VIDEO~\cite{emuvideo} all follow the image-conditioned T2V pipeline. They inject image control conditions into the T2V generation process through an additional Appearance-conditioned network or conditioned latent concatenation. Utilizing a reference image improves visual fidelity and reduces artifacts, allowing the model to focus more on learning video dynamics.

SVD~\cite{svd} validates the data scaling capability of the video diffusion model. It collects and labels hundreds of millions of video-text data, and releases the image2video video generation model, which has been utilized in many subsequent works.

\noindent$\bullet$ \textbf{Complex Dynamics Modeling}
The generation of Text-to-Video (T2V) encounters challenges in modeling complex dynamics, particularly regarding disruptions in action coherence. To address this, Dysen-VDM~\cite{Dysen-VDM} introduces a method that transforms textual information into dynamic scene graphs. Leveraging Large Language Model (LLM)~\cite{chatgpt}, Dysen-VDM~\cite{Dysen-VDM} identifies pivotal actions from the input text and arranges them chronologically, enriching scenes with pertinent descriptive details. Furthermore, the model benefits from in-context learning of LLM, endowing it with robust spatio-temporal modeling. This approach demonstrates remarkable superiority in the synthesis of complex actions.

VideoDirGPT~\cite{videodirectorgpt} also utilizes LLM to plan the generation of video content. For a given text input, it is expanded into a video plan through GPT-4~\cite{GPT4}, which includes scene descriptions, entities along with their layouts, and the distribution of entities within backgrounds. Subsequently, corresponding videos are generated by the model with explicit control over layouts. This approach demonstrates layout and motion control advantages for complex dynamic video generation.

\noindent$\bullet$ \textbf{Domain-specific T2V Generation} 
Video-Adapter~\cite{videoAdapter} introduces a novel setting by transferring pre-trained general T2V models to domain-specific T2V tasks. By decomposing the domain-specific video distribution into pretrained noise and a small training component, it substantially reduces the cost of transferring training. The efficacy of this approach is verified in T2V generation for Ego4D~\cite{ego4d} and Bridge Data~\cite{bridgedata} scenarios.

NUWA-XL~\cite{nuwaxl} employs a coarse-to-fine generative paradigm, facilitating parallel video generation. It initially employs global diffusion to generate key frames, followed by utilizing a local diffusion model to interpolate between two frames. This methodology enables the creation of lengthy videos spanning up to 3376 frames, thus establishing a benchmark for the generation of animations. This work focuses on the field of cartoon video generation, utilizing its techniques to produce cartoon videos lasting several minutes.

Text2Performer~\cite{Text2Performer} decomposes human-centric videos into appearance and motion representations. It first employs unsupervised training on natural human videos using a VQVAE~\cite{VQVAE} latent space to disentangle appearance and pose representations. Subsequently, it utilizes a continuous VQ-diffuser~\cite{gu2022vector, bond2022unleashing} to sample continuous pose embeddings. Finally, it employs a motion-aware masking strategy in the spatio-temporal domain on the pose embeddings to enhance temporal correlations.

\noindent$\bullet$ \textbf{Transformer-based T2V Generation} The emergence of  Sora~\cite{Sora} significantly boosts the popularity of video diffusion models, utilizing the Diffusion Transformer architecture to achieve substantial improvements in video stability and consistency.
W.A.L.T~\cite{walt} explores Transformer-based video diffusion models, using a causal encoder to compress images and videos into a unified space for joint training of the diffusion model. Additionally, it adopts a window attention architecture~\cite{liu2021swin} tailored for joint spatial and spatio-temporal generative modeling, while the employed cascading super-resolution network can generate stable videos at high resolutions.

CMD~\cite{cmd} proposes a video diffusion model that separates content and motion, first training an autoencoder to encode videos into content frames and motion latents. Then, it uses the Diffusion Transformer(DiT)~\cite{dit} architecture to generate motion vectors. This method enables the generation of high-resolution videos at a low computational cost.

Snap Video~\cite{snapvideo} proposes a modification to the EDM~\cite{k-diffusion} diffusion framework for generating high-resolution videos and treats images as high frame-rate videos to avoid image-video modality mismatches. It replaces the U-Net architecture with a transformer architecture~\cite{chen2023fit}
, scaling up to billions of parameters to demonstrate competitive results.

GenTron~\cite{gentron} adapts Diffusion Transformers (DiTs)~\cite{dit} from class to text conditioning, a process that involves thorough empirical exploration of the conditioning mechanism. Additionally, it proposes a motion-free guidance method to modulate the weight of motion information in the generated video.

\subsubsection{Training-free  T2V Diffusion Methods}
While former methods are all training-based T2V approaches that typically rely on extensive datasets like WebVid~\cite{webvid} or other video datasets~\cite{hd-vila, wang2023internvid}. Some recent researches~\cite{Text2video-zero,freebloom} aim at reducing heavy training costs by developing training-free T2V approaches, as will be introduced next.

Text2Video-Zero~\cite{Text2video-zero} utilizes the pre-trained T2I model Stable Diffusion~\cite{stablediffusion} for video synthesis. To maintain consistency across different frames, it performs a Cross-Attention mechanism between each frame and the first frame. Additionally, it enriches motion dynamics by modifying the sampling method of latent code. Moreover, this method can be combined with conditional generation and editing techniques such as ControlNet~\cite{controlnet} and Instruct-Pix2Pix~\cite{brooks2023instructpix2pix}, enabling the controlled generation of videos.

DirecT2V~\cite{DirecT2V} and Free-Bloom~\cite{freebloom}, on the other hand, introduce large language model (LLM)~\cite{llm1, GPT4} to generate frame-to-frame descriptions based on a single abstract user prompt. LLM directors are employed to breakdown user input into frame-level descriptions. Additionally, to maintain continuity between frames, DirecT2V~\cite{DirecT2V} uses a novel value mapping and dual-softmax filtering approach. Free-Bloom~\cite{freebloom} proposes a series of reverse process enhancements, which encompass joint noise sampling, step-aware attention shifting, and dual-path interpolation.
Experimental results demonstrate these modifications enhance the zero-shot video generation capabilities.

To handle intricate spatial-temporal prompts, LVD~\cite{LVD} first utilizes LLM~\cite{GPT4} to generate dynamic scene layouts and then employs these layouts to guide video generation. Its approach requires no training and guides video diffusion models by adjusting attention maps based on the layouts, enabling the generation of complex dynamic videos.

DiffSynth~\cite{diffsynth} introduces a latent in-iteration deflickering framework and a video deflickering algorithm to reduce flickering and produce coherent videos. It is applicable to various domains, including video stylization and 3D rendering.

{\color{black} TI2V-Zero~\cite{ti2v-zero} proposes a tuning-free method that incorporates image-conditioned control constraints to the T2V  method, enabling more controllable video generation. It employs a "repeat-and-slide" strategy along with DDPM~\cite{ddpm} inversion to inject the initial frame as a control condition into the video generation process, which is an effective approach for long video generation.}

\begin{figure*}[!t]
    \centering
    \includegraphics[width=0.9\textwidth]{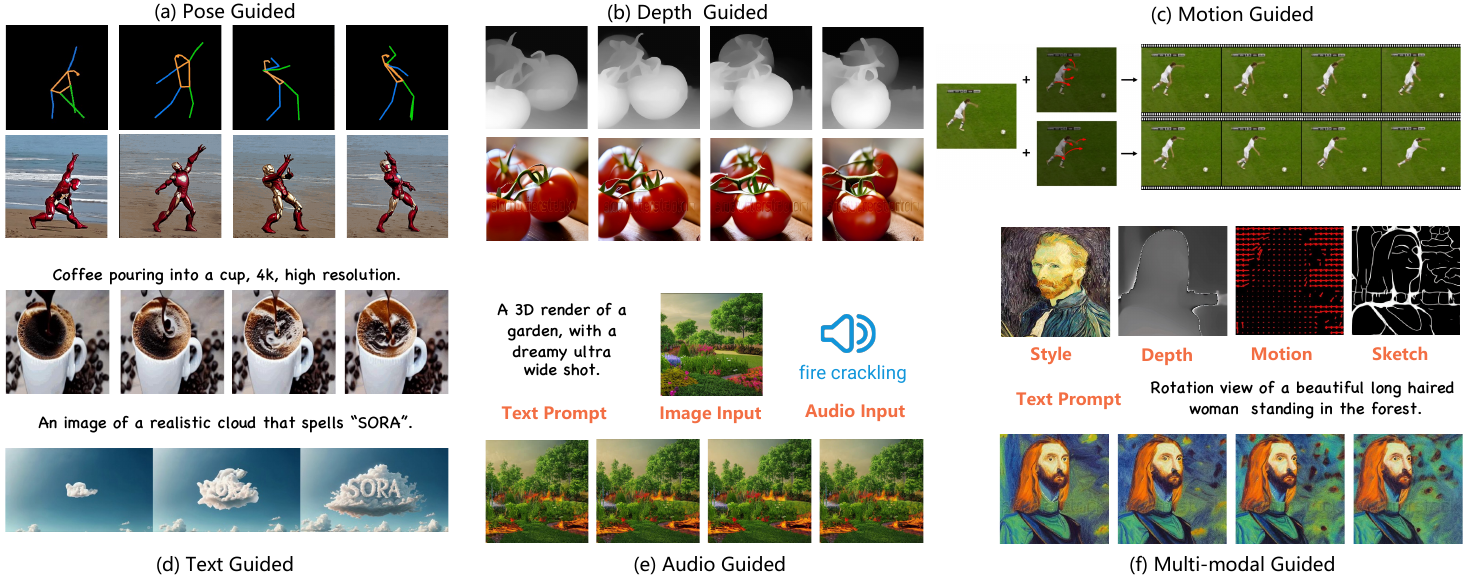}
    \vspace{-0.3cm}
    \caption{Conditional video generation results with (a) Pose Guided~\cite{FollowYourPose}, (b) Depth Guided~\cite{Make-Your-Video}, (c) Motion Guided~\cite{MCDiff}, (d) Text Guided~\cite{SimDA,Sora}, (e) Audio Guided~\cite{TPoS} and (f) Multi-modal Guided~\cite{videocomposer}.}
    \label{fig:conditon}
    \vspace{-0.6cm}
\end{figure*}

\subsection{Video Generation with other Conditions}

\label{Sec:modality-control}
Most of the previously introduced methods pertain to text-to-video generation. In this subsection, we focus on video generation conditioned on other modalities (\emph{e.g.} pose, sound and depth). We show the condition-controlled video generation examples in Fig.~\ref{fig:conditon}.

\subsubsection{Pose-guided Video Generation} Follow Your Pose~\cite{FollowYourPose} presents a video generation model driven by pose and text control. It employs a two-stage training process by utilizing image-pose pairs and pose-free videos. In the first stage, a T2I model is finetuned using (image, pose) pairs, enabling pose-controlled generation. In the second stage, the model leverages unlabeled videos to learn temporal modeling by incorporating temporal modules. This two-stage training imparts the model with both pose control and temporal modeling capabilities.

Dreampose~\cite{Dreampose} constructs a dual-path CLIP-VAE~\cite{clip} image encoder and adapter module to replace the original CLIP text encoder in LDM~\cite{stablediffusion} as the conditioning component. Given a single human image and a pose sequence, this study can generate a corresponding human pose video based on the provided pose information.

Dancing Avatar~\cite{DancingAvatar} focuses on synthesizing human dance videos. It utilizes a T2I model~\cite{stablediffusion} to generate each frame of the video in an auto-regressive manner. To ensure consistency throughout the entire video, a frame alignment module combined with insights from ChatGPT~\cite{chatgpt} is utilized to enhance coherence between adjacent frames. Additionally, it leverages OpenPose ControlNet~\cite{controlnet} to harness the ability to generate high-quality human body videos based on poses.

Disco~\cite{DisCo} addresses the novel problem of referring human dance generation using ControlNet~\cite{controlnet} for background control, Grounded-SAM~\cite{SAM} for foreground extraction, and OpenPose~\cite{openpose} for pose skeleton extraction. Large-scale image datasets~\cite{laion400m,coco,tiktok} are used for human attribute pre-training, creating a strong foundation for human-specific video generation tasks.

Animate Anyone~\cite{hu2024animateanyone} and MagicAnimate~\cite{xu2024magicanimate} use character images as reference frames and design a pose encoder to inject sequence pose information into the image-to-video generation, enabling characters to follow predefined poses for dancing. They demonstrate significant zero-shot capabilities with only a small training set, and the industrial applications prove the tremendous potential of video diffusion models.

PoseAnimate~\cite{poseanimate} proposes a zero-shot I2V framework for character animation. It introduces a Pose-Aware Control Module to control the pose, a Dual Consistency Attention Module to ensure temporal consistency, and a Mask-Guided Decoupling Module (MGDM) to refine distinct feature perception. Through these modules, the model demonstrates zero-shot pose animation capabilities.

\subsubsection{Motion-guided Video Generation} MCDiff~\cite{MCDiff} is the pioneer in considering motion as a condition for controlling video synthesis. The approach involves providing the first frame of a video along with a sequence of stroke motions. Initially, a flow completion model~\cite{wang2020deep} is utilized to predict dense video motion based on sparse stroke motion control. Subsequently, the model employs an auto-regressive approach using the dense motion map to predict subsequent frames, ultimately resulting in the synthesis of a complete video.

DragNUWA~\cite{DragNUWAFC} simultaneously introduces text, image, and trajectory information to provide fine-grained control over video content from semantic, spatial and temporal perspectives. To further address the lack of open-domain trajectory control in previous works, the authors proposed a Trajectory Sampler (TS) to enable open-domain control of arbitrary trajectories, a Multiscale Fusion (MF) to control trajectories in different granularities, and an Adaptive Training (AT) strategy to generate consistent video following trajectories.

{\color{black} MotionCtrl~\cite{motionctrl} proposes two control modules, one for camera motion and the other for object motion. It suggests methods for dataset collection tailored to these types of motion, and the trained control modules can be effectively integrated into various base models of video diffusion to enhance motion control.}

\subsubsection{Sound-guided Video Generation} AADiff~\cite{aadiff} introduces the concept of using audio and text together as conditions for video synthesis. The approach starts by separately encoding text and audio using dedicated encoders~\cite{clap}. Then, the similarity between the text and audio embeddings is computed, and the text token with the highest similarity is selected. This selected text token is used in a prompt2prompt~\cite{hertz2022prompt2prompt} fashion to edit frames. This approach enables the generation of audio-synchronized videos without requiring any additional training.

Generative Disco~\cite{generativeDisco} is designed for text-to-video generation aimed at music visualization. The system employs a pipeline that involves a large language model~\cite{GPT4} followed by a text-to-image model~\cite{stablediffusion} to achieve its goals.

TPoS~\cite{TPoS} integrates audio inputs with variable temporal semantics and magnitude, building upon the LDM~\cite{stablediffusion} to extend the utilization of audio modality in generative models. This approach outperforms widely-used audio-to-video benchmarks, as demonstrated by objective evaluations and user studies, highlighting its superior performance.

{\color{black}EMO~\cite{emo} proposes an audio-to-video framework for human talking generation, extracting audio features through an audio encoder and then injecting them into the diffusion model using audio attention. This allows for the creation of highly expressive and lifelike animations without the need for 3D facial information.}

\subsubsection{Image-guided Video Generation}
LaMD~\cite{lamd} first trains an autoencoder to separate motion information within videos. Then a diffusion-based motion generator is trained to generate video motion. Through this methodology, guided by motion, the model achieves the capability to generate high-quality perceptual videos given the first frame.

LFDM~\cite{LFDM} leverages conditional images and text for human-centric video generation. Firstly a latent flow auto-encoder is trained to reconstruct videos. Moreover, a flow predictor~\cite{wang2018video2video} can be employed in intermediary steps to predict flow motion. Subsequently, in the second stage, a diffusion model is trained with image, flow, and text prompts as conditions to generate coherent videos.

Generative Dynamics~\cite{generativeDynamics} presents an approach to modeling scene dynamics in image space. It extracts motion trajectories from real video sequences exhibiting natural motion. For a single image, the diffusion model, through a frequency-coordinated diffusion sampling process, predicts a long-term motion representation in the Fourier domain for each pixel. This representation can be converted into dense motion trajectories spanning the entire video. When combined with an image rendering module, it enables the transformation of static images into seamless looping dynamic videos, facilitating realistic user interactions with the depicted objects.

\subsubsection{Brain-guided Video Generation} MinD-Video~\cite{Cinematic} is the pioneering effort to explore video generation through continuous fMRI data. The approach begins by aligning MRI data with images and text using contrastive learning. Next, a trained MRI encoder replaces the CLIP text encoder as the input for conditioning. This is further enhanced through the design of a temporal attention module to model sequence dynamics. The resultant model is capable of reconstructing videos that possess precise semantics, motions, and scene dynamics, surpassing groundtruth performance and setting a new benchmark in this field.

\subsubsection{Depth-guided Video Generation} Make-Your-Video~\cite{Make-Your-Video} employs a novel approach for text-depth condition video generation. It integrates depth information as a conditioning factor by extracting it using MiDas~\cite{zeroshot-depth} during training. Additionally, the method introduces a causal attention mask to facilitate the synthesis of longer videos. Comparisons with state-of-the-art techniques demonstrate the method's superiority in controllable text-to-video generation, showcasing better quantitative and qualitative performance.

In Animate-A-Story~\cite{Animate-A-Story}, an innovative approach is introduced that divides video generation into two steps. The first step, Motion Structure Retrieval, involves retrieving the most relevant videos from a large video database based on a given text prompt~\cite{webvid}. Depth maps of these retrieved videos are obtained using offline depth estimation methods~\cite{zeroshot-depth}, which then serve as motion guidance. In the second step, Structure-Guided Text-to-Video Synthesis is employed to train a video generation model guided by the structural motion derived from the depth maps. Such a two-step approach enables the creation of personalized videos based on customized text descriptions. 

\subsubsection{Multi-modal guided Video Generation} VideoComposer~\cite{videocomposer}  focuses on  video generation conditioned on multi-modal, encompassing textual, spatial, and temporal conditions. Specifically, it introduces a Spatio-Temporal Condition encoder that allows flexible combinations of various conditions. This ultimately enables the incorporation of multiple modalities, such as sketch, mask, depth, and motion vectors. By harnessing control from multiple modalities, VideoComposer~\cite{videocomposer} achieves higher video quality and improved detail in the generated content.

MM-Diffusion~\cite{mm-diffusion} represents the inaugural endeavor in a joint audio-video generation. To realize the generation of multimodal content, it introduces a bifurcated architecture comprising two subnets tasked with video and audio generation, respectively. To ensure coherence between the outputs of these two subnets, a random-shift based attention block has been devised to establish interconnections. Beyond its capacity for unconditional audio-video generation, MM-Diffusion~\cite{mm-diffusion} also exhibits pronounced aptitude in effectuating video-to-audio translation.

MovieFactory~\cite{zhu2023moviefactory} is dedicated to applying the diffusion model to the generation of film-style videos. It leverages ChatGPT~\cite{raffel2020exploring,chatgpt} to elaborate on user-provided text, creating comprehensive sequential scripts for movie generation. In addition, an audio retrieval system has been devised to provide voice overs for videos. Through the aforementioned techniques, the realization of generating multi-modal audio-visual content is achieved.

CoDi~\cite{CoDi} presents a novel generative model that possesses the capability of creating diverse combinations of output modalities, encompassing language, images, videos, or audio, from varying combinations of input modalities. This is achieved by constructing a shared multimodal space, facilitating the generation of arbitrary modality combinations through the alignment of input and output spaces across diverse modalities.

NExT-GPT~\cite{next-gpt} presents an end-to-end, any-to-any multimodal LLM system. It integrates LLM~\cite{vicuna} with multimodal adapters and diverse diffusion decoders, enabling the system to perceive input in arbitrary combinations of text, images, videos, and audio, and generate corresponding output. 

VITRON~\cite{vitron}, utilizing an LLM backbone~\cite{vicuna}, incorporates various encoders for images and videos, enhancing the versatility of the model and supporting several tasks within a unified framework. Particularly, it achieves good results in both image-to-video and text-to-video settings, demonstrating the tremendous potential of combining LLMs with video diffusion models.

\subsection{Unconditional Video Generation}
\label{Sec:unconditional}
In this section, we delve into unconditional video generation. It refers to generating videos that belong to the specific domain without extra conditions. The focal points of these studies revolve around the design of video representations and the architecture of diffusion model networks.

\noindent$\bullet$ \textbf{U-Net based Generation}  
As one of the earliest works on unconditional video diffusion models and a significant baseline method, VIDM~\cite{VIDM} uses two streams: the content generation stream for video frame content generation and the motion stream to define video motion. By merging these streams, consistent videos are generated. Additionally, the authors employ Positional Group Normalization (PosGN)~\cite{improvedddpm} to enhance video continuity and explore combining Implicit Motion Condition (IMC) and PosGN to address the generation consistency of long videos.

Similar to LDM~\cite{stablediffusion}, PVDM~\cite{PVDM} first trains an auto-encoder to map pixels into a lower-dimensional latent space, followed by applying a diffusion denoising generative model in the latent space to synthesize videos. This approach reduces both training and inference costs while capable of maintaining satisfactory generation quality.

GD-VDM~\cite{gd-vdm} first generates depth map videos where scene and layout generation are prioritized whereas fine details and textures are abstracted away. Then, the generated depth maps are provided as a conditioning signal to further generate the remaining details of the video. This methodology retains superior detail generation capabilities and is particularly applicable to complex driving scene video generation tasks.

LEO~\cite{LEO} involves representing motion within the generation process through a sequence of flow maps, thereby inherently separating motion from appearance. It achieves human video generation through the combination of a flow-based image animator and a Latent Motion Diffusion Model. The former learns the reconstruction from flow maps to motion codes, while the latter captures motion priors to obtain motion codes. The synergy of these two methods enables effective learning of human video correlations. Furthermore, this approach can be extended to tasks such as infinite-length human video synthesis and content-preserving video editing.

{\color{black}\noindent$\bullet$ \textbf{Transformer-based Generation} 
Inspired by the success of DiT~\cite{dit} in image generation, VDT~\cite{vdt} and Latte~\cite{latte} aim to extend the diffusion transformer to the video generation tasks. They explore several spatio-temporal attention variants and conditional injection modules, while also validating the scaling capabilities of the video diffusion transformer.

\noindent$\bullet$ \textbf{Mamba-based Generation} Matten~\cite{matten} proposes a video diffusion model architecture that combines Mamba and Attention, exploring several spatio-temporal modeling variants based on Mamba and Attention, validating that Mamba can be applied to video diffusion models. DiM~\cite{DiM} explores a pure Mamba structure for image and video diffusion models, validating its scalability and ability to generate high-quality videos at a lower computational cost.
}

\subsection{Video Completion}
\label{Sec:video-completion}
Video completion constitutes a pivotal task within the realm of video generation. In the subsequent sections, we will delineate the distinct facets of video enhancement and restoration and video prediction.

\subsubsection{Video Enhancement and Restoration}

CaDM~\cite{CaDM} introduces a novel Neural-enhanced Video Streaming paradigm aimed at substantially diminishing streaming delivery bitrates, all the while maintaining a notably heightened restoration capability in contrast to prevailing methodologies. Primarily, the proposed CaDM~\cite{CaDM} approach improves the compression efficacy of the encoder through the concurrent reduction of frame resolution and color bit-depth in video streams. Furthermore, CaDM~\cite{CaDM} empowers the decoder with superior enhancement capabilities by imbuing the denoising diffusion restoration process with an awareness of the resolution-color conditions stipulated by the encoder.


LDMVFI~\cite{LDMVFI} is the first to use a conditional latent diffusion model for video frame interpolation (VFI). This approach introduces several innovative concepts, including a VFI-specific autoencoding network with efficient self-attention modules and deformable kernel-based frame synthesis techniques to improve performance.

VIDM~\cite{lookvidm} utilizes the pre-trained LDM~\cite{stablediffusion} for video inpainting. By providing a mask for first-person perspective videos, it leverages the  image completion capabilities of LDM to generate inpainted videos.

{\color{black}AVID~\cite{avid} first achieves fixed-length video in-painting by designing a motion module and structure guidance. Then, through the designed Temporal MultiDiffusion sampling pipeline with a middle-frame attention guidance mechanism, it extends in-painting to videos of arbitrary length.}

\subsubsection{Video Prediction}
Seer~\cite{Seer} is dedicated to the exploration of the text-guided video prediction task. It leverages the Latent Diffusion Model (LDM) as its foundational backbone. Through the integration of spatial-temporal attention within an auto-regressive framework, alongside the implementation of the Frame Sequential Text Decomposer module, Seer adeptly transfers the knowledge priors of Text-to-Image (T2I) models to the domain of video prediction. This migration has led to substantial performance enhancements, notably demonstrated on benchmarks~\cite{goyal2017somethingv2,bridgedata}.

FDM~\cite{FDM} introduces a novel hierarchy sampling scheme for the purpose of long video prediction tasks. Additionally, a new CARLA~\cite{carla} dataset is proposed. In comparison to auto-regressive methods, the proposed approach is not only more efficient but also yields superior generative outcomes.

MCVD~\cite{mcvd} employs a probabilistic conditional score-based denoising diffusion model for both unconditional generation and interpolation tasks. The introduced masking approach is capable of masking all past or future frames, thereby enabling the prediction of frames from either the past or the future. Additionally, it adopts an autoregressive approach to generate videos of variable lengths in a block-wise fashion. 

LGC-VD~\cite{LGCVD} introduces a Local-Global Context guided Video Diffusion model designed to encompass diverse perceptual conditions. LGC-VD employs a two-stage training approach and treats prediction errors as a form of data augmentation. This strategy effectively addresses prediction errors and notably reinforces stability in the context of long video prediction tasks.

RVD~\cite{RVD} adopts a diffusion model that utilizes the context vector of a convolutional Recurrent Neural Network as conditions to generate a residual, which is then added to a deterministic next-frame prediction. The authors demonstrate that employing residual prediction is more effective than directly predicting future frames. 

RaMViD~\cite{RaMViD} employs 3D convolutions to extend the image diffusion model into the realm of video tasks. It introduces a novel conditional training technique and utilizes a mask condition to extend its applicability to various completion tasks, including video prediction~\cite{BAIR,kinetics600}, infilling~\cite{BAIR}, and upsampling~\cite{ucf101, kinetics600}.

{\color{black}AID~\cite{aid} first proposed transferring a general video diffusion model to specific domain video prediction tasks. It introduces the DQFormer architecture, which injects the initial frame, user instructions, and multi-modal large model~\cite{llava} planning predictions into the video generation process. Additionally, it utilizes a lightweight adapter for efficient transfer SVD~\cite{svd} to robotics and first-person video generation.}

\begin{table*}[]
\centering
\scalebox{0.56}{
\begin{tabular}{@{}rccccccccccc@{}}
\toprule
\multirow{2}{*}{Method} & \multirow{2}{*}{Year} & \multirow{2}{*}{Training Data} & \multirow{2}{*}{\multirow{2}{*}{\begin{tabular}[c]{@{}c@{}}Extra\\ Dependency\end{tabular}}} & \multirow{2}{*}{Resolution} & \multirow{2}{*}{Params(B)} & \multicolumn{3}{c}{MSRVTT~\cite{msrvtt}} & \multicolumn{3}{c}{UCF-101~\cite{ucf101}} \\ \cmidrule(l){7-12} 
                        &                       &                                &                        &                             &                         & FID($\downarrow$)     & FVD($\downarrow$)    & CLIPSIM($\uparrow$)   & FID($\downarrow$)       & FVD($\downarrow$)      & IS($\uparrow$)      
\\ \hline
\rowcolor{gray!25}  \multicolumn{12}{c}{\bf Non-diffusion based method}                    \\ \hline

CogVideo~\cite{hong2022cogvideo}                & 2022                  & \cite{webvid}(5.4M)                   &  -                      & $256\times256$                          & 15.5                    & 23.59   & 1294  & 0.2631   & 179.00   & 701.59  & 25.27  \\
MMVG~\cite{mmvg}                    & 2023                  & \cite{webvid}(2.5M)                   &    -                    & $256\times256$                          & -                       & -        & -      & 0.2644   &   -       &   -      &  -      \\ \hline
\rowcolor{gray!25}  \multicolumn{12}{c}{\bf Diffusion based method}                    \\ \hline

LVDM~\cite{lvdm}                    & 2022                  & \cite{webvid}(2M)                     &   -                     & $256\times256$                         & 1.16                    & -       & 742   & 0.2381   & -         & 641.8   & -      \\

MagicVideo~\cite{magicvideo}              & 2022                  & \cite{webvid}(10M)                       & -                       & $256\times256$                          & -                       & -        & 998   &    -      & 145.00   & 699.00  &  -      \\
Make-A-Video~\cite{singer2022make}            & 2022                  & \cite{webvid, hd-vila}                    &  -                      & $256\times256$                          & 9.72                    & 13.17   &  -     & 0.3049   & -         & 367.23  & 33.00  \\
ED-T2V~\cite{edt2v}                  & 2023                  & \cite{webvid}(10M)                       &  -                      & $256\times256$                          & 1.30                    & -        &  -     & 0.2763   &  -        &   -      &  -      \\

InternVid~\cite{wang2023internvid}               & 2023                  & \cite{webvid}(10M) + 18M*                 &     -                   & $256\times256$                          & -                       & -        &  -     & 0.2951   &  60.25    & 616.51  & 21.04  \\
Video-LDM~\cite{videoLDM}               & 2023                  & \cite{webvid}(10M)                       &  -                      & $256\times256$                          & 4.20                    & -        &    -   & 0.2929   &   -       & 550.61  & 33.45  \\
VideoComposer~\cite{videocomposer}           & 2023                  & \cite{webvid}(10M)                       &   -                     & $256\times256$                          & 1.85                    &-         & 580   & 0.2932   & -         &  -       & -       \\
Latent-shift~\cite{latentshift}            & 2023                  & \cite{webvid}(10M)                    &  -                      & $256\times256$                          & 1.53                    & 15.23   & -     & 0.2773   & -         & -        &   -     \\
VideoFusion~\cite{videofusion}             & 2023                  & \cite{webvid}(10M)                    &  -                      & $256\times256$                          & 1.83                    &  -       & 581   & 0.2795   & 75.77    & 639.90  & 17.49  \\
Make-Your-Video~\cite{Make-Your-Video}         & 2023                  & \cite{webvid}(10M)                       & Depth Input            & $256\times256$                          & -                       & -        & -      & -         & -         & 330.49  &        \\
PYoCo~\cite{pyoco}                   & 2023                  & \cite{webvid} (22.5M)                        &   -                     & $256\times256$                          & -                       &  9.73    &   -    &   -       &  -        & 355.19  & 47.76  \\
CoDi~\cite{CoDi}                    & 2023                  & \cite{webvid,hd-vila}            &  -                      & $512\times512$                          & -                       &  -       &  -     & 0.2890   &  -        & -        & -       \\
NExT-GPT~\cite{next-gpt}    & 2023                  & \cite{webvid,hd-vila}            &  -                      & $320\times576$                          & 1.83                       & 13.04        & -      & 0.3085   &  -        &   -      &  -      \\
SimDA~\cite{SimDA}                   & 2023                  & \cite{webvid}(10M)                    &   -                     & $256\times256$                          & \bf 1.08                    &  -       &  456   & 0.2945   &  -        & -        &  -      \\

Dysen-VDM~\cite{Dysen-VDM}               & 2023                  & \cite{webvid}(10M)                    & ChatGPT                & $256\times256$                          & -                       & 12.64   &       & \bf 0.3204   &  -        &  325.42  & 35.57  \\
VideoFactory~\cite{videofactory}            & 2023                  & \cite{webvid, hd-vila}                   &   -                     & $256\times256$                          & 2.04                    &  -       & -      & 0.3005   & -         & 410.00  & -      \\
ModelScope~\cite{modelscope}            & 2023                  & \cite{webvid}(10M)                    &   -                     & $256\times256$                          & 1.70                    & 11.09        & 550      & 0.2930   &  -        & 410.00  &  -     \\
VideoGen~\cite{videogen}            & 2023                  & \cite{webvid}(10M)                    & Reference Image                       & $256\times256$                          & -                    & -       & -      & 0.3127   &  -        & 554.00  & \bf 71.61      \\
Animate-A-Story~\cite{Animate-A-Story}           & 2023                  & \cite{webvid}(10M)                    &  Depth Input                      & $256\times256$                          & -                  &-         &-       &-  & -         & 515.15  &       \\
VidRD~\cite{VidRD}           & 2023                  & \cite{webvid,kinetic,videolt}(5.3M*)                    &   -                     & $256\times256$                          & -                  &   -      &  -     & - &  -        & 363.19  &39.37       \\
LAVIE~\cite{lavie}               & 2023                  & \cite{webvid}(10M)+25M*                       & -                       & $320\times512$                          & 3.00                    & -        & -      & 0.2949   & -         & 526.30  & -  \\
VideoDirGPT~\cite{videodirectorgpt}               & 2023                  & \cite{webvid}(10M)                       &  GPT-4                      & $256\times256$                          & 1.92                    & 12.22        & 550      & 0.2860   &  -        & -  & -  \\
Show-1~\cite{show1}               & 2023                  & \cite{webvid}(10M)                       & -                      & $320\times576$                          &  -                   & 13.08        & 538      & 0.3072   & -         &394.46   &35.42   \\
Dynamicrafter~\cite{dynamicrafter}             & 2023                  & \cite{webvid}(10M)                    &  Reference Image                      & $256\times256$                          & -                    &  -       &  234   & -   & -    & 429.23  & -  \\
EMU-Video~\cite{emuvideo}             & 2023                  & 34M*                    &  Reference Image                      & $256\times256$                          & -                    &  -       & -   & -   & -    & 606.20  & 42.70  \\
PixelDance~\cite{pixeldance}             & 2023                  & \cite{webvid}(10M)+50W*                    &  Reference Image                      & $256\times256$                          & 1.50                    &  -       & 381   & 0.3125   & 49.36    & 242.82  & 42.10  \\
MicroCinema~\cite{microcinema}    & 2023                  & \cite{webvid}(10M)                    &  Reference Image                      & $256\times256$                          & 2.42                    &  -       & 377   & 0.2967   &-    & 342.86  & 37.46  \\
ART$\cdot$V~\cite{microcinema}    & 2023                  & \cite{webvid}(5M)                    &  Reference Image                      & $256\times256$                          & -                    &  -       & 291   & 0.2859   &-    & 315.69  & 50.34  \\
SVD~\cite{svd}    & 2023                  & 577M*                    &  Reference Image                      & $256\times384$                          & -                    &  -       & -   & -   &-    &  242.02  & -  \\
W.A.L.T~\cite{walt}    & 2023                  & 89M*                    &  -                      & $128\times128$                          & 3.00                    &  -       & 244.7   & -   &-    &  258.1  & 35.1   \\
CMD~\cite{cmd}    & 2023                  & \cite{webvid}(10M)                     &  Reference Image                      & $512\times1024$                          & 1.60                    &  -       & -   & 0.2894   &-    &  504.00  & -   \\
Snap Video~\cite{snapvideo}    & 2024                  & 238k hours                     &  -                     & $288\times512$                          & 3.90                    & \bf 9.35       & \bf 104.0   & 0.2793   &\bf 28.1    &  \bf 200.20  & 38.89   \\
\bottomrule
\end{tabular}
}
\caption{Zero-shot Text-to-Video generation comparison on MSR-VTT~\cite{msrvtt} and UCF-101~\cite{ucf101} dataset. We report the Fréchet Video Distance (FVD) scores,  CLIPSIM scores, Fréchet Image Distance (FID) and Inception Score (IS). The dataset marked with ``*" indicates the use of a self-collected dataset.}
\vspace{-0.8cm}
\label{Tab:zeroshot}
\end{table*}
\begin{table*}[]
\centering
\scalebox{0.66}{
\begin{tabular}{@{}rcccccccccc@{}}
\toprule
\multirow{2}{*}{Method} & \multirow{2}{*}{Year} & \multirow{2}{*}{Type} & \multirow{2}{*}{Resolution} & \multirow{2}{*}{Extra}   & \multicolumn{2}{c}{UCF-101~\cite{ucf101}} & \multicolumn{2}{c}{Taichi-HD~\cite{taichi-hd}} & \multicolumn{2}{c}{Time-lapse~\cite{time-lapse}} \\ \cmidrule(l){6-11} 
                        &                       &                             &      &                    & FVD($\downarrow$)           & IS($\uparrow$)          & FVD($\downarrow$)            & KVD($\downarrow$)           & FVD ($\downarrow$)           & KVD($\downarrow$)            \\ \midrule

MoCoGAN~\cite{tulyakov2018mocogan}                 & 2018   &GAN                    & $64\times64$                            & -                         &  -             & 12.42       &   -             & -             & 206.6          & -              \\
TGANv2~\cite{videoIS}                  & 2020  &GAN                & $128\times128$                         &  -                        & -             & 26.60       &  -              &  -            & -               &  -             \\
StyleGAN-V~\cite{skorokhodov2022stylegan}              & 2022   &GAN                    &   $256 \times256$                           &   -                       &  -             & 23.94       &  -              &  -            & 79.52          & -              \\
MoCoGAN-HD~\cite{mocoganhd}              & 2021  &GAN                 & $256 \times256$                        &   -                       & 700           & 33.95       & 144.7          & 25.4         & 183.6          & 13.9          \\
DIGAN~\cite{DIGAN}                   & 2022   &GAN                & $128\times 128$                         &  -                        & 577           & 32.70       & 128.1          & 20.6         & 114.6          & 6.8           \\
StyleInV~\cite{StyleInV}                   & 2023   &GAN                &  $256 \times256$                             &    -                      &  -          &   -    & 186.72               &  -            &  77.04              &   -            \\
MMVG~\cite{mmvg}                    & 2023  &VQGAN               & $128\times 128$                                &   -                       &    -           & 58.3        & 395            &   -           &   -             &  -             \\

VideoGPT~\cite{yan2021videogpt}                & 2021  &Autoregressive                 & $64\times64$                            &   -                       & -             & 24.69       & -               &   -           & 222.7          &  -             \\

CCVS~\cite{le2021ccvs}                    & 2021   &Autoregressive                    & $128\times 128$                             &  -                        & 386           & 24.47       & -               &  -            &   -             &   -            \\
TATS~\cite{TATS}                    & 2022  &Autoregressive                 & $128\times128$                         & Class Condition          & 278           & 79.28       & 94.6           & 9.8          & 132.6          & 5.7           \\

CogVideo~\cite{hong2022cogvideo}                & 2022 &Autoregressive                  & $160 \times160$                         & Pretrain+Class Condition & 626           & 50.46       &  -              & -             &  -              &  -             \\ \midrule       
VDM~\cite{vdm}                     & 2022   &Diffusion                & $64\times64$                          &  -                        & -             & 57.80       &  -              & -             & -               &   -            \\
LVDM~\cite{lvdm}                    & 2022   &Diffusion                & $256 \times256$                         &   -                       & 372           & 27.00            & 99             & 15.3             & 95.2           & \bf 3.9              \\

VIDM~\cite{VIDM}                    &  2022  &Diffusion                    & $256\times256$                         &   -                       & 294.7         &  -           & 121.9          &  -            & 57.4           & -              \\
LEO~\cite{LEO}                    &  2022  &Diffusion                    & $256\times256$                         & -                         & -         & -            & 122.7          & 20.49             & 57.4           &   -            \\
VideoFusion~\cite{videofusion}             & 2023  &Diffusion                 & $128\times128$                         &   -                       & 220           & 72.22       & \bf 56.4           & \bf 6.9          &   47.0           & 5.3           \\
PVDM~\cite{PVDM}                    & 2023    &Diffusion                   &  $256 \times256$                             &   -                       & 343.6         & 74.40       & -               & -             & 55.41          &  -             \\
VDT~\cite{vdt}                     & 2023   &Diffusion                & $64  \times64$                        &      -                    & 283.0         &  -           &  -              &  -            &  -              & -              \\
PYoCo~\cite{pyoco}                   & 2023   &Diffusion                &  $256 \times256$                             &     -                     & 310           & 60.01       &  -              & -             & -               &  -             \\

Dysen-VDM~\cite{Dysen-VDM}               & 2023    &Diffusion               &  $256 \times256$                             & ChatGPT                         & 255.42        & \bf 95.23       &  -              & -             &  -              &  -             \\
Latent-Shift~\cite{latentshift}            & 2022   &Diffusion                & $256 \times256$                         & Class Condition          & 360           & 92.72       & -               &  -            & -               &   -            \\
ED-T2V~\cite{edt2v}                  & 2023  &Diffusion                 & $256 \times256$                        & Class Condition          & 320           & 83.36       &  -              &  -            &  -              &   -            \\
Make-A-Video~\cite{singer2022make}            & 2023   &Diffusion                &$256\times256$                         & Pretrain+Class Condition & \bf 81.25         & 82.55       &  -              & -             &  -              &  -             \\
VideoGen~\cite{videogen}            & 2023   &Diffusion                &$256\times256$                         & Pretrain+Class Condition & 345       &  82.78       & -               &  -            &  -              &   -            \\
Matten~\cite{matten}            & 2024 &Diffusion                &$256\times256$                         & - & 210.61       & -      & 158.56               &  -            &  53.56              &   -            \\
DiM~\cite{DiM}            & 2024 &Diffusion                &$256\times256$                         & - & 206.83       & -      & -               &  -            &  -              &   -            \\
Latte~\cite{latte}            & 2024 &Diffusion                &$256\times256$                         & - & 333.61       & 73.31      &  97.09               &  -            & \bf 42.67              &   -            \\
AID~\cite{aid}  & 2024 &Diffusion                &$256\times256$                         & Pretrain+Class Condition & 102       & -      &  -               &  -            & -              &   -            \\
                             \bottomrule
\end{tabular}}
\caption{Finetuned video generated results of UCF-101~\cite{ucf101}, Taichi-HD~\cite{taichi-hd} and Time-lapse~\cite{time-lapse}. We report the FVD, IS and KVD scores evaluation metric of clips with 16 frames. We also report the resolution of each video frame for each evaluation result.}
\vspace{-0.7cm}
\label{Tab:unconditional_generation}
\end{table*}

\subsection{Benchmark Results}
\label{Sec:Benchmark}
This section systematically compares various video generation methods under zero-shot and finetuned settings. For each setting, we start by introducing their commonly used datasets. Subsequently, we state the detailed evaluation metrics utilized for each of the datasets. Finally, we present a comprehensive performances comparison of the methods.

\subsubsection{Zero-shot T2V Generation}
\noindent$\bullet$ \textbf{Datasets.}
General T2V methods, such as Make-A-Video~\cite{singer2022make} and VideoLDM~\cite{videoLDM}, are primarily evaluated on the MSRVTT~\cite{msrvtt} and UCF-101~\cite{ucf101} datasets in a zero-shot manner.  MSRVTT~\cite{msrvtt} is a video retrieval dataset, where each video clip is accompanied by approximately 20 natural sentences for description. Typically, the textual descriptions corresponding to the 2,990 video clips in its test set are utilized as prompts to produce the corresponding generated videos. UCF-101~\cite{ucf101} is an action recognition dataset with 101 action categories. In the context of T2V models, videos are typically generated based on the category names or manually set prompts corresponding to these action categories.

\noindent$\bullet$ \textbf{Evaluation Metrics.}
When evaluating under the zero-shot setting, it is common practice to assess video quality using FVD~\cite{fvd} and FID~\cite{fid} metrics on the MSRVTT~\cite{msrvtt} dataset. CLIPSIM~\cite{clip} is used to measure the alignment between text and video. For the UCF-101~\cite{ucf101} dataset, the typical evaluation metrics include Inception Score~\cite{videoIS}, FVD~\cite{fvd}, and FID~\cite{fid} to evaluate the quality of generated videos and their frames.

\noindent$\bullet$ \textbf{Results Comparison.}
In Table~\ref{Tab:zeroshot}, we present the zero-shot performance of current general T2V methods on MSRVTT~\cite{msrvtt} and UCF-101~\cite{ucf101}. We also provide information about their parameter number, training data, extra dependencies, and resolution. It can be observed that methods relying on ChatGPT~\cite{Dysen-VDM} or other input conditions~\cite{Make-Your-Video, Animate-A-Story} exhibit a significant advantage over others, and the utilization of additional data~\cite{singer2022make, wang2023internvid, videofactory} often leads to improved performance.

\subsubsection{Finetuned Video Generation}
\noindent$\bullet$ \textbf{Datasets.} Finetuned video generation methods refer to generating videos after fine-tuning on a specific dataset. This typically includes unconditional video generation and class conditional video generation. It primarily focuses on three specific datasets: UCF-101~\cite{ucf101}, Taichi-HD~\cite{taichi-hd}, and Time-lapse~\cite{time-lapse}. These datasets are associated with distinct domains: UCF-101 concentrates on human sports, Taichi-HD mainly comprises Tai Chi videos, and Time-lapse predominantly features time-lapse footage of the sky. Additionally, there are several other benchmarks available~\cite{kinetic, BAIR, cordts2016cityscapes}, but we choose these three as they are the most commonly used ones. 

\noindent$\bullet$ \textbf{Evaluation Metrics.}
In the evaluation of the Finetuned Video Generation task, commonly used metrics for the UCF-101~\cite{ucf101} dataset include IS~\cite{videoIS} (Inception Score) and FVD~\cite{fvd} (Fréchet Video Distance). For the Time-lapse~\cite{time-lapse} and Taichi-HD~\cite{taichi-hd} datasets, common evaluation metrics include FVD and KVD~\cite{kvd}.

\noindent$\bullet$ \textbf{Results Comparison.}
In Table~\ref{Tab:unconditional_generation}, we present the performance of current state-of-the-art methods fine-tuned on benchmark datasets. Similarly, further details regarding the method type, resolution, and extra dependencies are provided. It is evident that diffusion-based methods exhibit a significant advantage compared to traditional GANs~\cite{tulyakov2018mocogan,mocoganhd,DIGAN} and autoregressive Transformer~\cite{yan2021videogpt, hong2022cogvideo} methods. Furthermore, if there is a large-scale pretraining or class conditioning, the performance tends to be further enhanced.

\begin{figure*}[htbp]
	\centering
	\begin{minipage}{0.62\linewidth}
		\centering
            \tikzstyle{my-box}=[
    rectangle,
    draw=hidden-draw,
    rounded corners,
    text opacity=1,
    minimum height=1.5em,
    minimum width=5em,
    inner sep=2pt,
    align=center,
    fill opacity=.5,
    line width=0.8pt,
]
\tikzstyle{leaf}=[my-box, minimum height=1.5em,
    fill=hidden-pink!80, text=black, align=left,font=\scriptsize,
    inner xsep=2pt,
    inner ysep=4pt,
    line width=0.8pt,
]
    \centering
    \resizebox{1.0\textwidth}{!}{
        \begin{forest}
            forked edges,
            for tree={
                grow=east,
                reversed=true,
                anchor=base west,
                parent anchor=east,
                child anchor=west,
                base=left,
                font=\scriptsize,
                rectangle,
                draw=hidden-draw,
                rounded corners,
                align=left,
                minimum width=1em,
                edge+={darkgray, line width=1pt},
                s sep=3pt,
                inner xsep=2pt,
                inner ysep=3pt,
                line width=0.8pt,
                ver/.style={rotate=90, child anchor=north, parent anchor=south, anchor=center},
            },
            where level=1{text width=4.3em,font=\scriptsize,}{},
            where level=2{text width=4.5em,font=\scriptsize,}{},
            [
                Video Editing                (\S \ref{Sec:videoediting})
                , ver
                [
                    Text-guided
                    \\ (\S \ref{Sec:text-guided-video-editing})
                    [
                        Training-based
                        [
                            GEN-1~\cite{gen1}{,} Dreamix~\cite{dreamix}{,} TCVE~\cite{TCVE}{,} MagicEdit~\cite{magicedit}  \\  Control-A-Video~\cite{Control-A-Video}{,} MagicProp~\cite{magicprop}{,} \textcolor{black}{FlowVid~\cite{flowvid}}
                            , leaf, text width=17.5em
                        ]
                    ]
                    [
                        Training-free
                        [
                            TokenFlow~\cite{geyer2023tokenflow}{,} EVE~\cite{EVE}{,} VidEdit~\cite{couairon2023videdit}{,} FateZero~\cite{qi2023fatezero} \\ Rerender-A-Video~\cite{yang2023rerenderavideo}{,} 
                                  Pix2Video~\cite{ceylan2023pix2video}{,} MeDM~\cite{MeDM}{,}  
                            \\Ground-A-Video~\cite{groundavideo}{,}
                            Vid2Vid-Zero~\cite{vid2vid-zero}{,} InFusion~\cite{infusion}
                            \\
                            ControlVideo$_\text{1}$~\cite{controlvideo}{,}  Gen-L-Video~\cite{Gen-L-Video}{,}
                            FLATTEN~\cite{flatten}
                            , leaf, text width=17.5em
                        ]
                    ]
                    [
                        One-shot-tuned
                        [
                            SAVE~\cite{save}{,} StableVideo~\cite{chai2023stablevideo}{,}
                            Shape-aware TLVE~\cite{satlve}  \\
                            Edit-A-Video~\cite{shin2023editavideo}{,}  SinFusion~\cite{Sinfusion}{,} ControlVideo$_\text{2}$~\cite{controlvideoone-shot}\\ EI$^2$~\cite{ei2}{,} Tune-A-Video~\cite{tuneavideo}{,}                       
                             Video-P2P~\cite{liu2023videop2p}
                            , leaf, text width=17.5em
                        ]
                    ]
                ]
                [
                     Modality 
                    \\guided (\S \ref{Sec:modality-guided-video-editing})
                    [
                        Instruct-guided
                        [
                           Instruct-vid2vid\cite{qin2023instructvid2vid}{,} CSD~\cite{CSD}{,} 
                            \textcolor{black}{VIDiff~\cite{vidiff}{,} Fairy~\cite{fairy}}
                            , leaf, text width=17.5em
                        ]
                    ]
                    [
                        Sound-guided
                        [
                            Soundini~\cite{lee2023soundini}{,} SDVE~\cite{sdve}
                            , leaf, text width=8em
                        ]
                    ]
                    [
                        Motion-guided
                        [
                            VideoControlNet~\cite{videocontrolnet}{,} \textcolor{black}{MotionEditor~\cite{motioneditor}}
                            , leaf, text width=13em
                        ]
                    ]
                    [
                        Multi-Modal
                        [
                            Make-A-Protagonist~\cite{zhao2023makeaprotagonist} {,} 
                             CCEdit~\cite{ccedit} {,} \textcolor{black}{DreamVideo~\cite{dreamvideo}}
                            , leaf, text width=17.5em
                        ]
                    ]
                ]
                [
                    Domain 
                    \\Specific (\S \ref{Sec:domain-specific-video-editing})
                    [
                        Recolor\&Restyle, text width=5.2em
                        [
                            ColorDiffuser~\cite{colorDiffusers}{,} Style-A-Video~\cite{styleavideo}{,} \textcolor{black}{Diffutoon~\cite{diffutoon}}
                            , leaf, text width=16em
                        ]
                    ]
                    [
                        Human Video
                        [
                            Diffusion Video Autoencoders~\cite{diffusionvideoautoencoder}{,}   TGDM~\cite{tgdm} \\
                            Instruct-Video2Avatar~\cite{instruct-video2avatar}
                            , leaf, text width=15em
                        ]
                    ]
                ]
            ]
        \end{forest}
    }
    \caption{Taxonomy of Video Editing. Key aspects of Video Editing include General Text-guided Video Editing, Modality-guided Video Editing and Domain-specific Video Editing. }
    \label{taxo_of_mcot}
	\end{minipage}
         \hspace{0.02\textwidth} 
        \begin{minipage}{0.33\linewidth}
		\centering
            \tikzstyle{my-box}=[
    rectangle,
    draw=hidden-draw,
    rounded corners,
    text opacity=1,
    minimum height=1.5em,
    minimum width=5em,
    inner sep=2pt,
    align=center,
    fill opacity=.5,
    line width=0.8pt,
]
\tikzstyle{leaf}=[my-box, minimum height=1.5em,
    fill=hidden-pink!80, text=black, align=left,font=\footnotesize,
    inner xsep=2pt,
    inner ysep=4pt,
    line width=0.8pt,
]
    \centering
    \resizebox{1.0\linewidth}{!}{
        \begin{forest}
            forked edges,
            for tree={
                grow=east,
                reversed=true,
                anchor=base west,
                parent anchor=east,
                child anchor=west,
                base=left,
                font=\footnotesize,
                rectangle,
                draw=hidden-draw,
                rounded corners,
                align=left,
                minimum width=4em,
                edge+={darkgray, line width=1pt},
                s sep=3pt,
                inner xsep=2pt,
                inner ysep=3pt,
                line width=0.8pt,
                ver/.style={rotate=90, child anchor=north, parent anchor=south, anchor=center},
            },
            where level=1{text width=5.8em,font=\footnotesize,}{}, 
            [
                Video Understanding (\S \ref{Sec:video-understanding})
                , ver
                [
                    Action Detection \\ \& Segmentation
                    [
                        DiffTAD~\cite{difftad}{,} DiffAct~\cite{diffact}
                        , leaf, text width=10em
                    ]
                ]
                [
                    Video Anomaly \\ Detection
                    [
                        Diff-VAD~\cite{DiffVAD}{,} CMR~\cite{UVAD} \\ MoCoDAD~\cite{MoCoDAD}
                        , leaf, text width=10em
                    ]
                ]
                [
                    Text-Video \\ Retrieval
                    [
                        DiffusionRet~\cite{jin2023diffusionret} \\ MomentDiff~\cite{momentdiff} \\ DiffusionVMR~\cite{diffusionvmr}
                        , leaf, text width=7.5em
                    ]
                ]  
                [
                    Video \\ Captioning
                    [
                         RSFD~\cite{RSFD}
                        , leaf, text width=4em
                    ]
                ]   
                [
                    Video Object \\ Segmentation
                    [
                        Pix2Seq-D~\cite{pix2seq-D}
                        , leaf, text width=5.5em
                    ]
                ]     
                [
                    Video Pose \\ Estimation
                    [
                        DiffPose~\cite{feng2023diffpose}
                        , leaf, text width=5.5em
                    ]
                ]  
                [
                    Audio-Video \\ Separation
                    [
                        DAVIS~\cite{huang2023davis}
                        , leaf, text width=4.5em
                    ]
                ]
                [
                    Action \\ Recognition 
                    [
                        DDA~\cite{DDA} {,} GenRec~\cite{genrec}
                        , leaf, text width=9.3em
                    ]
                ]
                [
                    Video \\ Soundtracker 
                    [
                        LORIS~\cite{LORIS}
                        , leaf, text width=4.5em
                    ]
                ]
                [
                    Video Procedure \\ Planning 
                    [
                        PDPP~\cite{wang2023pdpp}
                        , leaf, text width=4.5em
                    ]
                ]
            ]
        \end{forest}
    }
    \caption{Taxonomy of diffusion-based Video Understanding. }
    \label{taxo_of_lavr}
		\label{图2}
	\end{minipage}
    \vspace{-0.2cm}
\end{figure*}


\section{Video Editing}
\label{Sec:videoediting}

With the development of diffusion models, there has been an exponential growth in the number of research studies in video editing.
As a consensus of many researches~\cite{dreamix, controlvideoone-shot, infusion, chai2023stablevideo}, video editing tasks should satisfy the following criteria: (1) fidelity: each frame should be consistent in content with the corresponding frame of the original video; (2) alignment: the output video should be aligned with the input control information; (3) quality: the generated video should be temporal consistent and in high quality. While a pre-trained image diffusion model can be utilized for video editing by processing frames individually, the lack of semantic consistency across frames renders editing a video frame by frame infeasible, making video editing a challenging task. In this section, we divide video editing into three categories: Text-guided video editing (Sec.~\ref{Sec:text-guided-video-editing}), Modality-guided video editing (Sec.~\ref{Sec:modality-guided-video-editing}), and Domain-specific video editing (Sec.~\ref{Sec:domain-specific-video-editing}). The taxonomy details of video editing are summarized in Fig.~\ref{taxo_of_mcot}.

\subsection{Text-guided Video Editing}
\label{Sec:text-guided-video-editing}

In text-guided video editing, the user provides an input video and a text prompt which describes the desired attributes of the resulting video. However, unlike image editing, text-guided video editing represents new challenges of frame consistency and temporal modeling. In general, there are two main ways for text-based video editing: (1) training a T2V diffusion model on a large-scale text-video pairs dataset and (2) extending the pre-trained T2I diffusion models for video editing. The latter garnered more interest due to the fact that large-scale text-video datasets are hard to acquire, and training a T2V model is computationally expensive. To capture motion in videos, various temporal modules are introduced to T2I models. Nonetheless, methods inflating T2I models suffer from two critical issues: \textit{Temporal inconsistency}, where the edited video exhibits flickering in vision across frame, and \textit{Semantic disparity}, where videos are not altered in accordance with the semantics of given text prompts. Several studies address the problems from different perspectives.

\subsubsection{Training-based Methods}
The training-based approach refers to the method of training on a large-scale video-text dataset, enabling it to serve as a general video editing model.

GEN-1~\cite{gen1} proposes a structure and content-aware model that fully controls temporal, content, and structural consistency. This model introduces temporal layers into a pre-trained T2I model and trains it jointly on images and videos, achieving real-time control over temporal consistency.

Dreamix~\cite{dreamix} proposes two innovations: starting generation with a low-resolution version of the original video and fine-tuning the model on the original video. Additionally, they propose mixed fine-tuning with full temporal attention and temporal attention masking, enhancing motion editability.

TCVE~\cite{TCVE} introduces a cohesive spatial-temporal modeling unit to connect the temporal U-Net and the pre-trained T2I U-Net, which effectively captures the temporal coherence of input videos.

Control-A-Video~\cite{Control-A-Video} is based on a pre-trained T2I diffusion model, incorporating a spatio-temporal self-attention module and trainable temporal layers. Additionally, they propose a first-frame conditioning strategy, allowing it to produce videos of any length using an auto-regressive method.

MagicEdit~\cite{magicedit} separates the learning of content, structure, and motion in different frameworks, achieving high fidelity and temporal coherence.

MagicProp~\cite{magicprop} improves video editing by separating appearance editing from motion-aware propagation. It edits a reference frame and then uses a diffusion model to generate each frame based on the previous frame, target depth, and reference appearance.

Unlike previous methods use flow as hard constraints, FlowVid\cite{flowvid} considers the potential imperfections in flow estimation. In this way, they include depth map as additional spatial condition, along with the temporal flow condition, enabling consistent and flexible video editing.

\subsubsection{Training-free Methods}
Training-free approach involves utilizing pre-trained T2I or T2V models and adapting them for video editing tasks in a zero-shot manner. Compared to training-based methods, training-free methods require no heavy training cost. However, they may suffer a few potential drawbacks. First of all, videos edited in a zero-shot manner may produce spatio-temporal distortion and inconsistency. Furthermore, methods utilizing T2V models might still incur high training and inference costs. We briefly examine the techniques used to address these issues.

TokenFlow~\cite{geyer2023tokenflow} demonstrates that consistency in edited videos can be achieved by enforcing consistency in the diffusion feature space. Specifically, this is accomplished by sampling keyframes, jointly editing them, and propagating the features from the keyframes to all other frames based on the correspondences provided by the original video features. This process explicitly maintains consistency and a fine-grained shared representation of the original video features.

VidEdit~\cite{couairon2023videdit} combines atlas-based~\cite{bar2022text2live} and pre-trained T2I~\cite{stablediffusion} models, which not only exhibit high temporal consistency but also provide object-level control over video content appearance. The method involves decomposing videos into layered neural atlases with a semantically unified representation of content, and then applying a pre-trained, text-driven image diffusion model for zero-shot atlas editing. Concurrently, it preserves structure in atlas space by encoding both temporal appearance and spatial placement.

Rerender-A-Video~\cite{yang2023rerenderavideo} employs hierarchical cross-frame constraints to enforce temporal consistency. It uses optical flow to apply dense cross-frame constraints, with the previously rendered frame as a low-level reference and the first rendered frame as an anchor to maintain consistency in style, shape, texture, and color.

To address the issues of heavy costs in atlas learning~\cite{bar2022text2live} and per-video tuning~\cite{tuneavideo}, FateZero stores comprehensive attention maps at every stage of the inversion process to maintain superior motion and structural information. Additionally, it incorporates spatial-temporal blocks to enhance visual consistency.

Vid2Vid-Zero~\cite{vid2vid-zero} utilizes a null-text inversion~\cite{Nulltext} module to align text with video, a spatial regularization module for video-to-video fidelity, and a cross-frame modeling module for temporal consistency. Similar to FateZero~\cite{qi2023fatezero}, it also incorporates a spatial-temporal attention module.

Pix2Video~\cite{ceylan2023pix2video} initially utilizes a pre-trained structure-guided T2I model to conduct text-guided edits on an anchor frame, ensuring the generated image remains true to the edit prompt. Subsequently, they progressively propagate alterations to future frames using self-attention feature injection, maintaining temporal coherence.

InFusion~\cite{infusion} consists of two components. First, it integrates features from the residual block in decoder layers and attention features into the denoising pipeline for the editing prompt, showcasing its zero-shot editing capability. Second, it merges attention for edited and unedited concepts using mask extraction from cross-attention maps, ensuring consistency.

ControlVideo$_\text{1}$~\cite{controlvideo} directly adopts the weights from ControlNet~\cite{controlnet}, extending self-attention with fully cross-frame interaction to achieve high-quality and consistency. To manage long-video editing tasks, it implements a hierarchical sampler that divides the long video into short clips and attains global coherence by conditioning on pairs of keyframes.

EVE~\cite{EVE} proposes two strategies to reinforce temporal consistency: \textit{Depth Map Guidance} to locate spatial layouts and motion trajectories of moving objects as well as \textit{Frame-Align Attention} which forces the model to place attention on both previous and current frames. 

MeDM~\cite{MeDM} utilizes explicit optical flows to establish a pragmatic encoding of pixel correspondences across video frames, thus maintaining temporal consistency. Furthermore, they iteratively align noisy pixels across video frames using the provided temporal correspondence guidance derived from optical flows.

Gen-L-Video~\cite{Gen-L-Video} addresses long video editing by treating them as overlapping short segments. Using Temporal Co-Denoising methods, it adapts existing short video editing models~\cite{tuneavideo, lvdm, ceylan2023pix2video} to edit videos with hundreds of frames while preserving consistency.

FLATTEN~\cite{flatten} incorporates optical flow into the attention mechanism of the diffusion model. The proposed Flow-guided attention allows patches from different frames to align on the same flow path within the attention module, enabling mutual attention and enhancing video editing consistency.

\subsubsection{One-shot-tuned Methods}
One-shot tuned method entails fine-tuning a pre-trained T2I model using a specific video instance, enabling the generation of videos with similar motion or content. While it requires extra training expenses, these approaches provide greater editing flexibility compared to training-free methods.

SinFusion~\cite{Sinfusion} pioneers one-shot-tuned diffusion-based models that learn motions from a single input video using only a few frames. Its backbone is a fully convolutional DDPM~\cite{ddpm} network, allowing it to generate images of any size.

SAVE~\cite{save} finetunes the spectral shift of the parameter space such that the underlying motion concept as well as content information in the input video is learned. Also, it proposes a spectral shift regularizer to restrict the changes. 

Edit-A-Video~\cite{shin2023editavideo} contains two stages: the first stage inflates a pre-trained T2I model to the T2V model and finetunes it using a single \textless text, video\textgreater~pair while the second stage is the conventional diffusion and denoising process. A key observation is that edited videos often suffer from background inconsistency. To address such an issue, they propose a masking method called \textit{sparse-causal blending}, which automatically generates a mask to approximate the edited region.

Tune-A-Video~\cite{tuneavideo} leverages a sparse spatio-temporal attention mechanism that only visits the first and the former video frames, together with an efficient tuning strategy that only updates the projection matrices in the attention blocks. Furthermore, it seeks structural guidance from input video at inference time to make up for the lack of motion consistency.  

Instead of using a T2I model, Video-P2P~\cite{liu2023videop2p} alters it into a Text-to-set model (T2S) by replacing self-attentions with frame-attentions, which yields a model that generates a set of semantically-consistent images. Furthermore, they use a decoupled-guidance strategy to improve the robustness to the change of prompts.

ControlVideo$_\text{2}$~\cite{controlvideoone-shot} mainly focuses on improving attention modules in the diffusion model and ControlNet~\cite{controlnet}. They transform the original spatial self-attention into key-frame attention, which aligns all frames with a selected one. Additionally, they incorporate temporal attention modules to preserve consistency.

Shape-aware TLVE~\cite{satlve} utilizes the T2I model and handles shape changes by propagating the deformation field between the input and edited keyframe to all frames.

EI$^2$~\cite{ei2} makes two key innovations: the Shift-restricted Temporal Attention Module (STAM) to restrict newly introduced parameters in the Temporal Attention module, resolving the semantic disparity, as well as the Fine-coarse Frame Attention Module (FFAM) for temporal consistency, which leverages the information on the temporal dimension by sampling along the spatial dimension. Combining these techniques, they create a T2V diffusion model.

StableVideo~\cite{chai2023stablevideo} designs an inter-frame propagation mechanism on top of the existing T2I model and an aggregation network to generate the edited atlases from the key frames, thus achieving temporal and spatial consistency.

\subsection{Other Modality-guided Video Editing}
Most of the methods introduced previously focus on text-guided video editing. In this subsection, we will focus on video editing guided by other modalities (\emph{e.g.}, Instruct and Sound).

\label{Sec:modality-guided-video-editing}

\subsubsection{Instruct-guided Video Editing}
Instruct-guided video editing aims to generate video based on the given input video and instructions. 
Due to the lack of video-instruction datasets, InstructVid2Vid~\cite{qin2023instructvid2vid} leverages the combined use of ChatGPT, BLIP~\cite{blip2}, and Tune-A-Video~\cite{tuneavideo} to acquire input videos, instructions and edited videos triplets at a relatively low cost. During training, they propose the Frame Difference Loss, guiding the model to generate temporal consistent frames.
CSD~\cite{CSD} first uses Stein variational gradient descent (SVGD), where multiple samples share their knowledge distilled from diffusion models to accomplish inter-sample consistency. Then, they combine Collaborative Score Distillation (CSD) with Instruct-Pix2Pix~\cite{brooks2023instructpix2pix} to achieve coherent editing of multiple images with instruction. \textcolor{black}{VIDiff~\cite{vidiff} is based on a multimodal instruction-guided editing diffusion model, unifying tasks including video editing, video recoloring, video object segmentation, and low-level tasks. It demonstrates the tremendous potential of video diffusion models and the feasibility of applying a unified architecture to various tasks.} \textcolor{black}{Fairy\cite{fairy} samples anchor frames where the features extracted from them can be propagated to frames with high similarity with them, allowing for parallel processing among different frame groups. Hence, it ensures temporal consistency by sharing global features as well as reduces the memory requirement.}

\subsubsection{Sound-guided Video Editing}
The goal of sound-guided video editing is to make visual changes consistent with the sound in the targeted region. 
To achieve this goal, Soundini~\cite{lee2023soundini} presents local sound guidance and optical flow guidance for diffusion sampling. Specifically, the audio encoder makes sound latent representation semantically consistent with the latent image representation.
Based on a diffusion model, SDVE~\cite{sdve} introduces a feature concatenation mechanism for temporal coherence. They further condition the network on speech by feeding spectral feature embeddings with the noise signal throughout the residual layers. 

\subsubsection{Motion-guided Video Editing}
Inspired by the video coding process, VideoControlNet~\cite{videocontrolnet} combines a diffusion model with ControlNet~\cite{controlnet}. It designates the first frame as the I-frame and divides the remaining frames into groups of pictures (GoP). The last frame of each GoP is set as the P-frame, while others are B-frames. For an input video, the model generates the I-frame using the diffusion model and ControlNet. P-frames are then generated through the motion-guided P-frame generation module (MgPG), leveraging optical flow information. Finally, B-frames are interpolated based on the reference I/P-frames and motion information, avoiding the time-consuming diffusion model. \textcolor{black}{MotionEditor\cite{motioneditor} features a two-branch architecture (\emph{i.e.} a reconstruction branch and an editing branch), which complements ControlNet\cite{controlnet} by seamlessly enforcing temporal motion correspondence, enabling high-fidelity editing and temporal consistency.}

\subsubsection{Multi-Modal Video Editing}
Make-A-Protagonist~\cite{zhao2023makeaprotagonist} introduces a multi-modal conditioned video editing framework to alter the protagonist. It employs BLIP-2~\cite{blip2} for video captioning, CLIP Vision Model~\cite{clip} and DALLE-2 Prior~\cite{dalle2} for visual and textual clue encoding, and ControlNet~\cite{controlnet} for video consistency. During inference, a mask-guided denoising sampling technique combines these elements to achieve annotation-free video editing.

CCEdit~\cite{ccedit} decouples video structure and appearance for controllable and creative video editing. It preserves the video structure using the foundational ControlNet~\cite{controlnet} while allowing appearance editing through text prompts, personalized model weights, and customized center frames. 

\textcolor{black}{DreamVideo\cite{dreamvideo} disentangles the personalized video editing task into two stages, subject learning and motion learning. They utilize a light-weight adapter to capture the appearance of given subject through texture inversion and another adapter to model target motion pattern, reducing the complexity of optimization and allowing for more flexible customization.}

\subsection{Domain-specific Video Editing}
In this subsection, we will provide a brief overview of several video editing techniques tailored for specific domains, starting with video recoloring and video style transfer methods in Sec.~\ref{Sec:Recolor-video-editing}, followed by several video editing methods designed for human-centric videos in Sec.~\ref{Sec:human-video-editing}.
\label{Sec:domain-specific-video-editing}

\subsubsection{Recolor \& Restyle}
\label{Sec:Recolor-video-editing}

\noindent$\bullet$ \textbf{Recolor} Video colorization aims to infer realistic and consistent colors for grayscale frames, balancing temporal, spatial, semantic consistency, color richness, and faithfulness. ColorDiffuser~\cite{colorDiffusers}, built on a pre-trained T2I model, introduces Color Propagation Attention to replace optical flow and an Alternated Sampling Strategy to capture spatio-temporal relationships between adjacent frames.

\noindent$\bullet$ \textbf{Restyle} Style-A-Video~\cite{styleavideo} designs a combined way of control conditions: text for style guidance, video frames for content guidance, and attention maps for detail guidance. Notably, the work features zero-shot, namely, no additional per-video training or fine-tuning is required. Diffutoon\cite{diffutoon} solve the toon shading task by dividing it into four sub-problems: stylization, consistency enhancement, structure guidance and colorization. They utilize Stable Diffusion\cite{stablediffusion} for stylization while ControlNet\cite{controlnet} is used for both outline-based generation and video coloring.

\subsubsection{Human Video Editing}
\label{Sec:human-video-editing}

Diffusion Video Autoencoders~\cite{diffusionvideoautoencoder} proposes a diffusion video autoencoder that extracts a single time-invariant feature (identity) and per-frame time-varient features (motion and background) from a given video and further manipulates the single invariant feature for the desired attribute, which enables temporal-consistent editing and efficient computing. 
Instruct-Video2Avatar~\cite{instruct-video2avatar} takes in a talking head video and an editing instruction and outputs an edited version of 3D neural head avatar. They simultaneously leverage~\cite{brooks2023instructpix2pix} for image editing, 
EbSynth~\cite{jamriska2018ebsynth} for video stylization, and INSTA~\cite{INSTA} for a photo-realistic 3D neural head avatar.
TGDM~\cite{tgdm} adopts the zero-shot CLIP-guided model to achieve flexible emotion control. Furthermore, they propose a pipeline based on the multi-conditional diffusion model to afford complex texture and identity transfer.

\section{Video Understanding}
\label{Sec:video-understanding}
In addition to its application in generative tasks, such as video generation and editing, diffusion model has also been explored in video understanding tasks 
such as video temporal segmentation~\cite{difftad, diffact}, video anomaly detection~\cite{DiffVAD, UVAD}, text-video retrieval~\cite{jin2023diffusionret, momentdiff}, \emph{etc.}, as will be introduced in this section. The taxonomy details of video understanding are summarized in Fig.~\ref{taxo_of_lavr}.

\noindent$\bullet$ \textbf{Temporal Action Detection\& Segmentation}
\label{Sec:detection-segmentation}
Inspired by DiffusionDet~\cite{diffusiondet}, 
DiffTAD~\cite{difftad} explores the application of diffusion models to the task of temporal action detection. 
This involves diffusing ground truth proposals of long videos and subsequently learning the denoising process, which is done by introducing a specialized temporal location query within the DETR~\cite{detr} architecture. Notably, the approach achieves state-of-the-art performance results on benchmarks such as ActivityNet~\cite{ANet-caption} and THUMOS~\cite{thumos}.

Similarly, DiffAct~\cite{diffact} addresses the task of temporal action segmentation using a comparable approach, where action segments are iteratively generated from random noise with input video features as conditions. The effectiveness of the proposed method is validated on widely-used benchmarks, including GTEA~\cite{GTEA}, 50Salads~\cite{50Salads}, and Breakfast~\cite{Breakfast}.

\noindent$\bullet$ \textbf{Video Anomaly Detection}
\label{Sec:video-anomaly-detection}
Dedicated to unsupervised video anomaly detection, Diff-VAD~\cite{DiffVAD} and CMR~\cite{UVAD} harness the reconstruction capability of the diffusion model to identify anomalous videos, as high reconstruction error typically indicates abnormality. Experiments conducted on two large-scale benchmarks~\cite{realanomaly,shanghaiTec} demonstrate the effectiveness of such a paradigm, consequently significantly improving performance compared to prior research.

MoCoDAD~\cite{MoCoDAD} focuses on skeleton-based video anomaly detection. The method applies the diffusion model to generate diverse and plausible future motions based on past actions of individuals. By statistically aggregating future patterns, anomalies are detected when a generated set of actions deviates from actual future trends.

\noindent$\bullet$ \textbf{Text-Video Retrieval}
\label{Sec:text-video retrieval}
DiffusionRet~\cite{jin2023diffusionret} approaches retrieval as generating a joint distribution $p(candidates, query)$ from noise. It combines generative and contrastive losses to train the generator and feature extractor, respectively, merging generative and discriminative techniques. This approach excels in open-domain scenarios, showing strong generalization ability.

MomentDiff~\cite{momentdiff} and DiffusionVMR~\cite{diffusionvmr} tackle video moment retrieval by converting time intervals into random noise and learning to denoise them back to their original states. This method trains the model to map random positions to actual intervals, improving the accuracy of localizing video segments based on textual descriptions.

\noindent$\bullet$ \textbf{Video Captioning}
\label{Sec:video-caption}
RSFD~\cite{RSFD} examines the frequently neglected long-tail problem in video captioning. It presents a new Refined Semantic enhancement approach for Frequency Diffusion (RSFD), which improves captioning by constantly recognizing the linguistic representation of infrequent tokens. This allows the model to comprehend the semantics of low-frequency tokens, resulting in enhanced caption generation.

\noindent$\bullet$ \textbf{Video Object Segmentation}
\label{Sec:video-object-segmentation}
Pix2Seq-D~\cite{pix2seq-D} redefines panoramic segmentation as a discrete data generation problem. It employs a diffusion model based on analog bits~\cite{Analogbits} to model panoptic masks, utilizing a versatile architecture and loss function. Furthermore, Pix2Seq-D~\cite{pix2seq-D} can model videos by incorporating predictions from previous frames, which enables the automatic learning of object instance tracking and video object segmentation.

\noindent$\bullet$ \textbf{Video Pose Estimation}
\label{Sec:video-pose-estimation}
DiffPose~\cite{feng2023diffpose} tackles video-based human pose estimation by treating it as a conditional heatmap generation task. It uses a Spatio-Temporal representation learner to aggregate features across frames and a multi-scale feature interaction mechanism to refine keypoint representations by establishing correlations across different scales.

\noindent$\bullet$ \textbf{Audio-Video Separation}
\label{Sec:audio-video-separation}
DAVIS~\cite{huang2023davis} tackles the audio-visual sound source separation task using a generative approach. The model employs a diffusion process to generate separated magnitudes from Gaussian noise, conditioned on the audio mixture and visual content.

\noindent$\bullet$ \textbf{Action Recognition}
\label{Sec:action-recognition}
DDA~\cite{DDA} enhances skeleton-based human action recognition by using diffusion-based data augmentation to generate high-quality, diverse action sequences. Experiments demonstrate the method's advantages in naturalness and diversity and its effectiveness when applied to existing action recognition models.

GenRec~\cite{genrec} explores applying spatiotemporal priors of video diffusion model to recognition tasks. It uses the pretrained SVD~\cite{svd} as the backbone and adds a classification head on top of the generative model, enabling the model to support both generation and classification tasks. Experimental results demonstrate that the two tasks can complement each other, achieving strong performance.

\noindent$\bullet$ \textbf{Video SoundTracker}
\label{Sec:soundtracker}
LORIS~\cite{LORIS} focuses on generating music soundtracks that synchronize with rhythmic visual cues. The system utilizes a latent conditional diffusion probabilistic model for waveform synthesis. Moreover, it incorporates context-aware conditioning encoders to account for temporal information, facilitating long-term waveform generation. The authors have also broadened the applicability of the model to various sports scenarios and is capable of producing long-term soundtracks with exceptional musical quality and rhythmic correspondence.

\noindent$\bullet$ \textbf{Video Procedure Planning }
\label{Sec:videoprocedure}
PDPP~\cite{wang2023pdpp} addresses procedure planning in instructional videos using a diffusion model to represent the distribution of the entire intermediate action sequence, transforming planning into a sampling process. The method employs a diffusion-based U-Net model for precise conditional guidance from initial and final observations, improving the learning and sampling of action sequences from the distribution.

\section{Challenges and Future Trends}
\label{Sec:future}

\begin{figure*}[!t]
    \centering
    \includegraphics[width=0.9\textwidth]{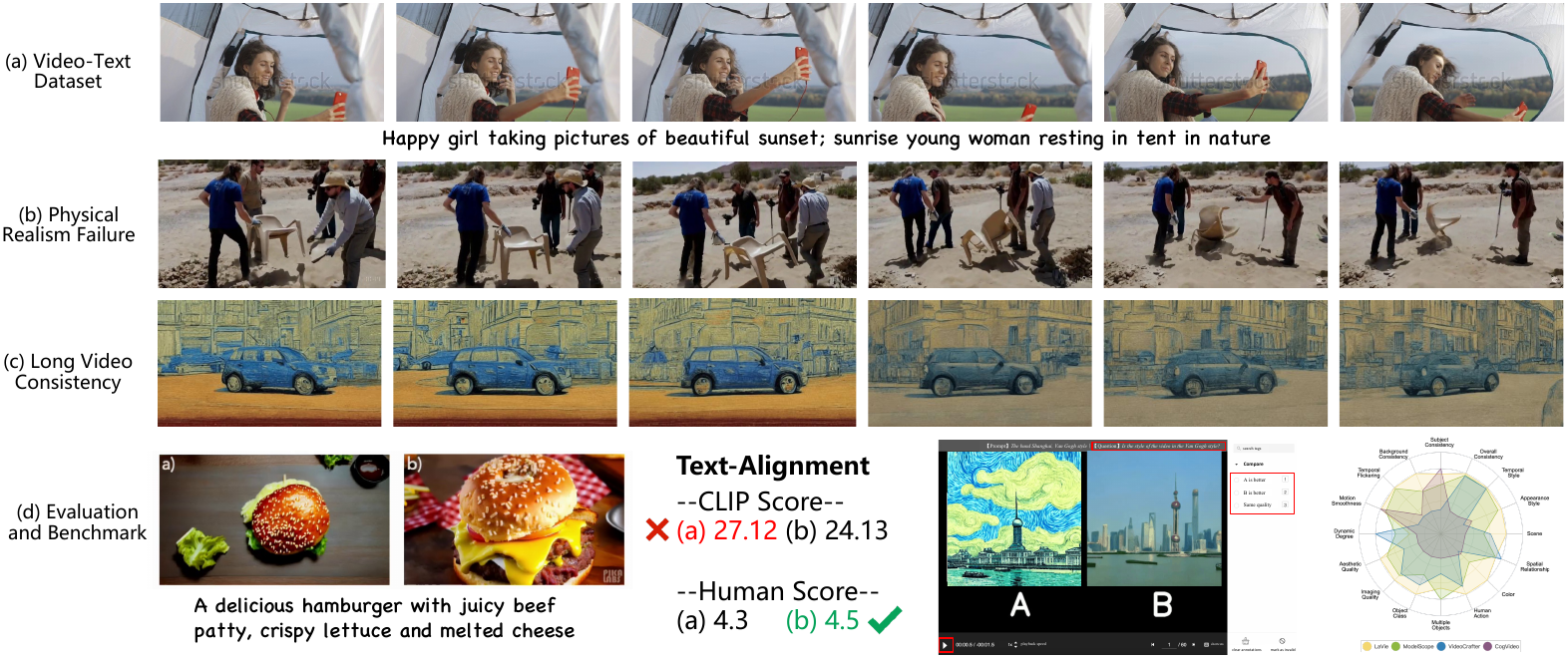}
    \vspace{-0.3cm}
    \caption{\textcolor{black}{\textbf{Challenges and future trends.} (a) Lack of high-quality video-text dataset~\cite{webvid}, (b) Physical Realism failure~\cite{Sora}, (c) Long video consistency failure~\cite{vidiff}, (d) Reliable Evaluation~\cite{bettermetric} and Benchmark~\cite{vbench}.}}
    \label{fig:future}
    \vspace{-0.6cm}
\end{figure*}

Despite the fact that diffusion-based methods have achieved significant advances in video generation, editing and understanding, there are still certain open problems worthy of exploration. In this section, we summarize the current challenges and potential future trends. We have also provided qualitative results in Fig.~\ref{fig:future} to illustrate the current limitations and future trends.

\noindent$\bullet$ \textbf{Collecting Large-scale Video-Text Datasets} The substantial achievements in Text-to-Image synthesis primarily stemmed from the availability of billions of high-quality (text, image) pairs. However, the commonly used datasets for Text-to-Video (T2V) tasks are relatively small in scale and gathering equally extensive datasets for video content is a considerably challenging endeavor. For example, the WebVid dataset~\cite{webvid} contains only 10 million instances and has a significant drawback of its limited visual quality, with a low resolution of 360P, further compounded by  watermark artifacts.
Despite ongoing efforts to develop new methods for obtaining datasets~\cite{videofactory,webvid,VidRD,chen2024panda}, there remains a pressing need to enhance dataset scale, annotation accuracy, and video quality.
The successes of SVD~\cite{svd} and Sora~\cite{Sora} demonstrate the effectiveness of scaling datasets. However, since they used private datasets, high-quality open-source datasets are crucial for video generation research.

\noindent$\bullet$ \textbf{Physical Realism and Long Video Generation} Even the most advanced video diffusion generation models, such as Sora~\cite{Sora}, exhibit some limitations in accurately describing complex scenes, showing inconsistencies in the portrayal of physical principles in videos of complex scenes, such as incorrect simulations of the rigid structure of chairs and unrealistic physical interactions. Moreover, most video generation models~\cite{videoLDM, singer2022make, SimDA} currently can only produce videos shorter than 10 seconds. Commonly used autoregressive methods~\cite{FlexiFilm} for generating long videos suffer from error accumulation, resulting in poorer quality in later frames. Besides, multi-stage coarse-to-fine methods~\cite{nuwaxl} for long video generation can be complex and time-consuming. Future efforts should aim to explore more physically accurate and realistic video generation, as well as consistency and stability in long video synthesis.

\noindent$\bullet$ \textbf{Efficient Training and Inference}
The heavy training cost associated with T2V models presents a significant challenge, with some tasks necessitating the use of hundreds of GPUs~\cite{videoLDM, VidRD}. Despite the efforts by methods~\cite{SimDA} to reduce training cost, both the magnitude of dataset and temporal complexity remain a critical concern. 
\textcolor{black}{With the rise of Sora~\cite{Sora}, the high training cost of training from scratch has become a significant challenge in video generation. More efficient compression of video representations~\cite{omnitokenizer}, exploration of effective spatiotemporal modeling methods~\cite{SimDA, matten}, and acceleration of training and inference times~\cite{videolcm, zhang2023adadiff} are important research directions.}

\noindent$\bullet$ \textbf{Benchmark and Evaluation Methods}
Although benchmarks~\cite{msrvtt,ucf101,vbench,aigcbench, fetv,evalcrafter} and evaluation methods~\cite{fvd,clip} for open-domain video generation exist, they are relatively limited in scope, as is demonstrated in~\cite{chivileva2023measuring}. Due to the absence of ground truth for the generated videos in Text-to-Video (T2V) generation, existing metrics such as Fréchet Video Distance (FVD)~\cite{fvd} and Inception Score (IS)~\cite{videoIS} primarily emphasize the disparities between generated and real video distributions. This makes it challenging to have a comprehensive evaluation metric that accurately reflects video generation quality. Currently, there is considerable reliance on user AB testing and subjective scoring, which is labor-intensive and potentially biased due to subjectivity. Constructing more tailored evaluation benchmarks~\cite{vbench, evalcrafter} and metrics~\cite{stream,contentbias, bettermetric} in the future is also a meaningful avenue of research.

\noindent$\bullet$ \textbf{More Controllable Video Editing}
The task of video editing has evolved alongside the development of video generation. Although existing video editing models have achieved impressive results in video style transfer~\cite{geyer2023tokenflow, yang2023rerenderavideo}, there are still limitations in certain tasks. For instance, previous video editing methods often exhibit noticeable temporal inconsistencies when controlling object replacement~\cite{tuneavideo}. Most video editing methods rely on detailed text descriptions, which limits their control and generality~\cite{vidiff, qin2023instructvid2vid}. Additionally, there are limitations in current methods for multi-object editing, motion editing~\cite{motioneditor}, and long video editing~\cite{Gen-L-Video}. As base models for video generation continue to develop~\cite{svd,Sora}, the future research trend will be towards more controllable video generation models with stronger generality and multi-modal capabilities.

\section{Conclusion}
\label{Sec:conclusion}
This survey offered an in-depth exploration of the latest developments in the era of AIGC (AI-generated Content) with a focus on video diffusion models. To the best of our knowledge, this is the first work of its kind. We provided a comprehensive overview of the
fundamental concepts 
of the diffusion process, popular benchmark datasets, and commonly used evaluation metrics.
Building upon this foundation, we comprehensively reviewed over 100 different
works focusing on the task of video generation, editing and understanding, and categorized them according to their technical perspectives and research objectives. 
Furthermore, in the experimental section, we meticulously described the experimental setups and conducted a fair comparative analysis across various benchmark datasets. In the end, we put forth several research directions for the future of video diffusion models.

\noindent\textbf{Acknowledge} This work was supported in part by National Natural Science Foundation of China (No. 62032006).

\vspace{-0.3cm}

{
\bibliographystyle{abbrv}

\bibliography{egbib}
}

\end{document}